%% file: main.tex
\pdfoutput=1

\documentclass[11pt]{article}

\PassOptionsToPackage{table,xcdraw,svgnames}{xcolor}
\usepackage[preprint]{acl}

\usepackage{times}
\usepackage{latexsym}

\usepackage[T1]{fontenc}

\usepackage[utf8]{inputenc}

\usepackage{microtype}

\usepackage{inconsolata}

\usepackage{graphicx}
\usepackage{tikz}
\usetikzlibrary{shapes.geometric}
\usepackage{array,multirow}
\usepackage{amssymb}
\usepackage{fixltx2e}

\usepackage{pifont}
\usepackage{tabularx}
\usepackage{booktabs}
\usepackage{xstring}    %
\usepackage{xspace}     %
\usepackage{graphicx}   %

\usepackage{hyperref}
\usepackage[most]{tcolorbox}

\definecolor{tableheadcolor}{HTML}{E8EAF6}
\definecolor{tablerowcolor}{HTML}{F8F9FA}
\definecolor{tableframecolor}{HTML}{D1D9E4}
\definecolor{filledstarcolor}{HTML}{FFCA28}
\definecolor{emptystarcolor}{HTML}{CFD8DC}
\definecolor{negativescalecolor}{HTML}{EF9A9A}
\definecolor{positivescalecolor}{HTML}{A5D6A7}
\newcommand{\filledstar}{\textcolor{filledstarcolor}{\ding{108}}}
\newcommand{\emptystar}{\textcolor{emptystarcolor}{\ding{109}}}

\newcommand{\rating}[1]{%
   \ifcase#1\relax%
     \emptystar\emptystar\emptystar\emptystar\emptystar\or%
     \filledstar\emptystar\emptystar\emptystar\emptystar\or%
     \filledstar\filledstar\emptystar\emptystar\emptystar\or%
     \filledstar\filledstar\filledstar\emptystar\emptystar\or%
     \filledstar\filledstar\filledstar\filledstar\emptystar\or%
     \filledstar\filledstar\filledstar\filledstar\filledstar%
   \fi%
 }

\usepackage{enumitem}
\usepackage{tcolorbox}
\tcbuselibrary{listingsutf8}

\definecolor{lightbluepastel}{RGB}{173,216,230}

\newtcolorbox{questionbox}[1][]{
  colframe=lightbluepastel,
  colback=lightbluepastel!20,
  boxrule=1pt,
  arc=4pt,
  auto outer arc,
  left=6pt,
  right=6pt,
  top=6pt,
  bottom=6pt,
  fonttitle=\bfseries,
  title=#1,
  breakable
}

\newcounter{question}[section]

\newenvironment{question}[1][]
  {\refstepcounter{question}\begin{questionbox}[#1]}
  {\end{questionbox}}

\newenvironment{choices}
  {\begin{enumerate}[label=\Alph*., wide=0pt, leftmargin=*]}
  {\end{enumerate}}

\newcommand{\choice}{\item}

\newcommand{\correctchoice}[1]{\item {\color{green!60!black}#1}}

\newcommand{\cetvel}{\textsc{Cetvel}}
\newcommand{\pergel}{\textsc{Cetvel}}

\definecolor{CBblue}{HTML}{005AB5}
\definecolor{CBorange}{HTML}{FFC20A}
\definecolor{CBred}{HTML}{DC3220}

\newcommand{\redsquare}{\tikz[baseline=-0.5ex]{\node[shape=rectangle, fill=CBred, draw, inner sep=0.8mm] {};}}
\newcommand{\bluesquare}{\tikz[baseline=-0.5ex]{\node[shape=rectangle, fill=CBblue, draw, inner sep=0.8mm] {};}}
\newcommand{\orangesquare}{\tikz[baseline=-0.5ex]{\node[shape=rectangle, fill=CBorange, draw, inner sep=0.8mm] {};}} 

\newcommand{\reddot}{\tikz[baseline=-0.5ex]{\node[shape=circle, fill=CBred, draw, inner sep=0.6mm] {};}}
\newcommand{\bluedot}{\tikz[baseline=-0.5ex]{\node[shape=circle, fill=CBblue, draw, inner sep=0.6mm] {};}}
\newcommand{\orangedot}{\tikz[baseline=-0.5ex]{\node[shape=circle, fill=CBorange, draw, inner sep=0.6mm] {};}}

\usepackage{tikz}
\newcommand{\DashedLineWithText}[2][6]{%
  \begin{tikzpicture}
    \draw[dashed] (0,0) -- (#1,0) node[midway, fill=white, inner sep=2pt] {#2};
  \end{tikzpicture}%
}

\newcommand{\pergelicon}[0]{\includegraphics[width=.05\textwidth]{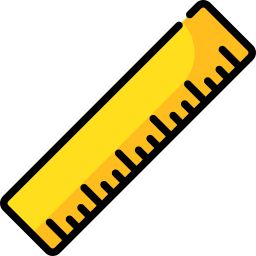}}

\newcommand{\huggingface}[1]{%
  \StrBehind{#1}{https://huggingface.co/}[\hfPathTemp]%
  \xspace
  \mbox{\raisebox{-0.8ex}{\includegraphics[width=1.5em, height=1.5em]{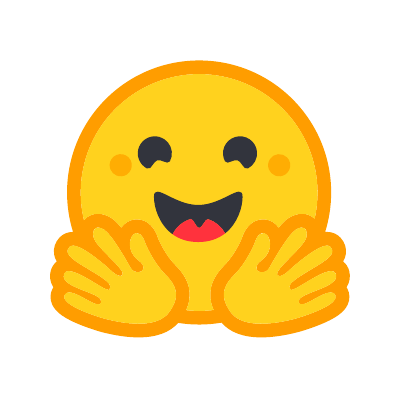}}\hspace{0.2em}}%
  \href{#1}{\texttt{\nolinkurl{\hfPathTemp}}}%
}

\newcommand{\fngithub}[1]{%
  \StrBehind{#1}{https://github.com/}[\githubPathTemp]%
  \mbox{%
    \xspace
    \raisebox{-0.1ex}{\includegraphics[width=0.8em, height=0.8em]{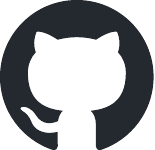}}%
    \hspace{0.2em}%
    \href{#1}{\footnotesize\texttt{\nolinkurl{\githubPathTemp}}}%
  }%
}

\title{\pergelicon\pergel: A Unified Benchmark for Evaluating Language Understanding, Generation  and Cultural Capacity of LLMs for Turkish}

\author{Abrek Er$^{1}$\textsuperscript{*} \ \ Ilker Kesen$^{3}$\textsuperscript{*} \ \ Gözde Gül Şahin$^{1,2}$ \ \ Aykut Erdem$^{1,2}$\\
$^{1}$ KUIS AI Center \ \ $^{2}$ Department of Computer Engineering, Koç University \\
$^{3}$Department of Computer Science, University of Copenhagen \\
\textbf{\textsuperscript{*}}Equal Contribution \ \
\texttt{aber@ku.edu.tr}}

\begin{document}
\maketitle
\input{sections/sec_abstract}
\input{sections/sec_introduction}

\input{sections/sec_related_work}
\input{sections/sec_datasets}
\input{sections/sec_experimental_setup}
\input{sections/sec_results}
\input{sections/sec_conclusion}
\bibliography{custom}

\appendix

\section{AI Assistant Usage}
Within this work, we only used AI Assistants for writing purposes.
We mainly used chatbots for refining our initial writing, i.e., proof-reading, improving clarity and coherence.
We did not use them for generating textual content based on instructions.

\section{Complete Results}
\label{sec:appendix_results}
This appendix subsection includes overall and task-specific results for all models tested within \cetvel. 

\clearpage
\onecolumn
\subsection{Overall Results}
\label{sec:overall_results}
\input{tables/results1_py}

\clearpage
\subsection{Grammatical Error Correction Results}
\label{sec:gec}
\input{tables/gec}

\clearpage
\subsection{Multiple Choice Question Answering Results}
\label{sec:mcqa}
\input{tables/mcqa}

\clearpage
\subsection{Machine Translation Results}
\label{sec:mt}
\input{tables/mt}

\clearpage
\subsection{Natural Language Inference Results}
\label{sec:nli}
\input{tables/nli}

\clearpage
\subsection{Open Ended Question Answering Results}
\label{sec:qa}
\input{tables/qa}

\clearpage
\subsection{Summarization Results}
\label{sec:sum}
\input{tables/sum}

\clearpage
\subsection{Text Classification Results}
\label{sec:tc}
\input{tables/tc}
\twocolumn

\section{Task Samples}
\label{sec:appendix_samples}
This appendix section includes sample instances for all tasks \& datasets included within \cetvel.

\input{sections/sec_appendix_samples}

\input{tables/datasets}

\end{document}

%% file: sections/sec_abstract.tex
\begin{abstract}
We introduce \cetvel, a comprehensive benchmark designed to evaluate large language models (LLMs) in Turkish. Existing Turkish benchmarks often lack either task diversity or culturally relevant content, or both. \cetvel~addresses these gaps by combining a broad range of both discriminative and generative tasks ensuring content that reflects the linguistic and cultural richness of Turkish language. \cetvel~covers 23 tasks grouped into seven categories, including tasks such as grammatical error correction, machine translation, and question answering rooted in Turkish history and idiomatic language. We evaluate 33 open-weight LLMs (up to 70B parameters) covering different model families and instruction paradigms. Our experiments reveal that Turkish-centric instruction-tuned models generally underperform relative to multilingual or general-purpose models (e.g. Llama 3 and Mistral), despite being tailored for the language. Moreover, we show that tasks such as grammatical error correction and extractive question answering are particularly discriminative in differentiating model capabilities. \cetvel~offers a comprehensive and culturally grounded evaluation suite for advancing the development and assessment of LLMs in Turkish.
\end{abstract}

%% file: sections/sec_introduction.tex
\section{Introduction}
\label{sec:introduction}
Large language models (LLMs) have recently achieved remarkable performance on widely used English-centric benchmarks such as (Super)GLUE \citep{wang2018glue,wang2019superglue} and MMLU \citep{hendryckstest2021}. Their success across a broad spectrum of tasks and domains \citep{jiang2023mistral7b,touvron2023llama,yang2024qwen2} has spurred the development of evaluation suites in languages beyond English \cite{park2021klue,elmadany-etal-2023-orca,nielsen-2023-scandeval}. In this work, we extend these efforts to Turkish by introducing \cetvel\footnote{\cetvel{} means \textit{ruler} in Turkish, i.e. a rectangular shaped object used for measuring the distance between two points.}
, a comprehensive benchmark designed to evaluate LLMs across a diverse set of natural language processing (NLP) tasks, with a particular emphasis on cultural and linguistic relevance to Turkish.

Existing Turkish NLP benchmarks typically suffer from one or both of the following limitations: insufficient task diversity and a lack of culturally relevant content. \cetvel~addresses both shortcomings. First, it provides broad task coverage, extending well beyond the multiple-choice question answering (MCQA) format predominant in recent Turkish benchmarks \citep{yuksel2024turkishmmlu,bayram2024setting,openllm-Turkish-leaderboard}. Specifically, \cetvel~includes 23 tasks grouped into seven categories: Text Classification (TC), Multiple Choice Question Answering (MCQA), Extractive Question Answering (QA), Grammatical Correction (GC), Machine Translation (MT), Summarization (SUM), and Natural Language Inference (NLI).

\begin{figure*}[h!]
\includegraphics[width=1.0 \textwidth]{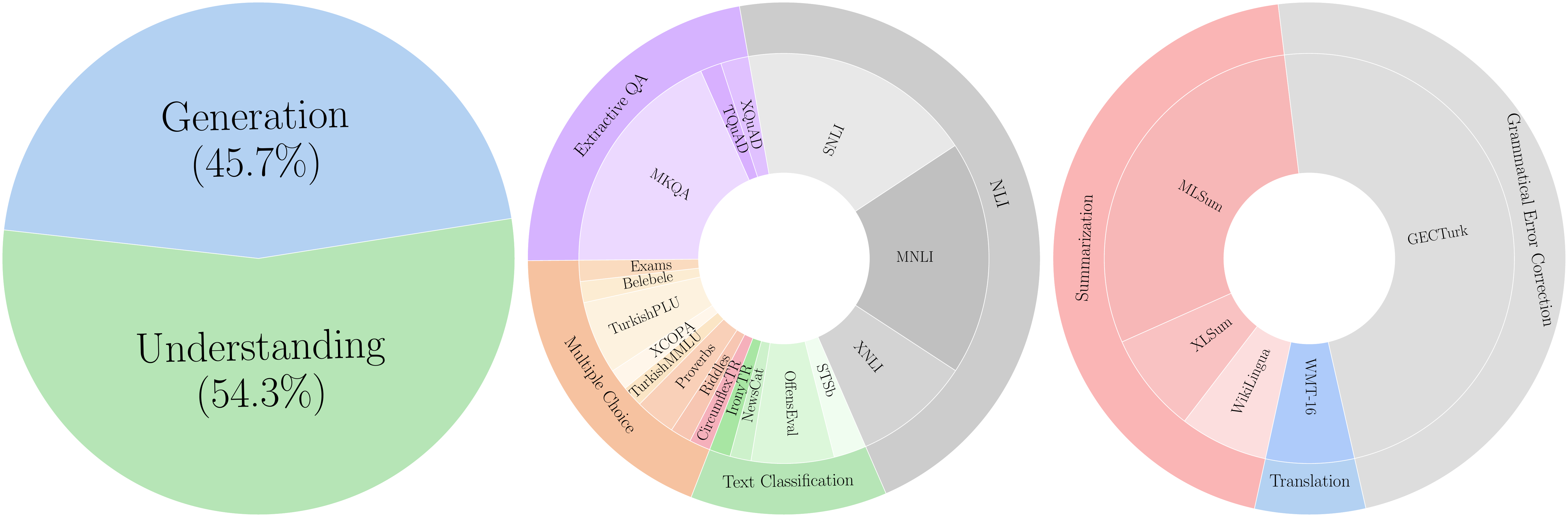}
    \caption{\textbf{Task taxonomy in the \cetvel~benchmark.} The leftmost pie chart illustrates the overall distribution of tasks across two primary categories: \textbf{language understanding} and \textbf{language generation}. The middle chart details the subtypes within the understanding category, including extractive question answering, multiple-choice QA, text classification, and natural language inference, along with their associated datasets. The rightmost chart breaks down the generation tasks into three subtypes: summarization, machine translation, and grammatical error correction.}
    \label{fig:data_dist}
\end{figure*}

Second, \cetvel~prioritizes content deeply rooted in Turkish language and culture, an aspect often missing in multilingual or machine-translated benchmarks, which tend to reflect Western cultural biases \citep{singh2024globalmmluunderstandingaddressing,acikgoz-etal-2024-bridging}. To counter this, \cetvel~includes tasks based on grammatical error correction, figurative language processing, and extractive QA centered on Turkish and Islamic history. We also introduce a novel circumflex-based word sense disambiguation task\footnote{In Turkish, \textit{hala} means aunt, whereas \textit{hâlâ} means still.}, further enriching the benchmark’s linguistic specificity.

We evaluate 33 open-weight LLMs on \cetvel, spanning a broad range of parameter scales (1B to 70B) and model families (e.g., Llama 3, Qwen2.5), including both general-purpose/multilingual and Turkish-specific models.
Among all models, Llama 3 variants consistently deliver the strongest overall performance within their respective size categories.
However, more importantly, our results show that most instruction-tuned LLMs specifically developed for Turkish do not outperform general-purpose models such as Llama 3 variants and Mistral.
Our findings suggest that Turkish-centric LLMs can benefit from improved instruction-tuning, continued pretraining, and more rigorous validation strategies.
Nonetheless, we find that there exists some exceptions:
Cere-Llama-3-8B achieves the best performance on grammatical error correction and extractive question answering about Turkish and Islamic history, even outmatching the 70 billion parameter model Llama-3.3-70B-Instruct.

Additionally, to better understand task-level variability, we assess the informativeness of each task using Gini coefficient-based analysis. Our findings indicate that grammatical error correction, machine translation, and extractive QA are particularly effective in differentiating model capabilities, positioning these tasks as highly valuable resources for benchmarking LLMs in Turkish.

\textbf{Our contributions are as follows:}
\begin{itemize}
\item We present \cetvel\footnote{Code and data: \fngithub{https://github.com/KUIS-AI/cetvel}}, a new Turkish LLM benchmark that combines broad task diversity with high linguistic and cultural relevance.
\item We evaluate 33 open-weight LLMs spanning multiple families, language specializations, and parameter scales (up to 70B).
\item Our results reveal that most Turkish-centric models do not outperform general-purpose LLMs such as Llama 3 and Mistral at comparable scales. 
\item We exceptionally find that Cere-Llama-3-8B excels among all other Turkish-centric LLMs, even surpassing a 70 billion parameter model on grammatical error correction and extractive QA about Turkish and Islam history.
\item We identify grammatical correction, machine translation, and extractive QA as the most informative tasks for evaluating Turkish LLMs.
\end{itemize}

%% file: sections/sec_related_work.tex
\section{Related Work}
\label{sec:related_work}

\subsection{LLM Benchmarks}
Early benchmark suites such as GLUE \citep{wang2018glue} and SuperGLUE \citep{wang2019superglue} have been pivotal in evaluating English-centric language understanding for smaller-scale models like BERT \citep{devlin-etal-2019-bert} and RoBERTa \citep{liu2019robertarobustlyoptimizedbert}. As LLMs have evolved \citep{dubey2024llama3herdmodels,yang2024qwen2}, more challenging benchmarks have emerged, targeting skills such as commonsense reasoning \citep{bisk2020piqa}, mathematical problem-solving \citep{cobbe2021gsm8k}, coding proficiency \citep{evalplus,evalperf}, and domain-specific knowledge (e.g., scientific QA) \citep{hendrycks2020measuring,rein2024gpqa}. Instruction-tuned models are further assessed through alignment and safety benchmarks \citep{Bai_2024}. Inspired by GLUE and SuperGLUE, \cetvel~brings a similar unification of tasks and datasets but with a specific focus on evaluating LLMs in Turkish.

\subsection{Multilingual Benchmarks}
Multilingual benchmarks such as XTREME \citep{hu2020xtreme}, XTREME-R \citep{ruder2021xtreme}, and XGLUE \citep{liang2020xglue} include Turkish among other languages, covering a variety of tasks like question answering (QA), natural language inference (NLI), machine translation (MT), and named entity recognition (NER). However, these benchmarks typically provide only one or two datasets per task, limiting their comprehensiveness. The MEGA benchmark \citep{ahuja-etal-2023-mega} extends multilingual evaluation by focusing on generative LLMs, featuring 16 datasets, including XQuAD~\citep{Artetxe:etal:2019}, MLQA~\citep{lewis2019mlqa}, XLSum~\citep{hasan-etal-2021-xl}, and WikiANN~\citep{rahimi-etal-2019-massively}, across 70 languages and multiple evaluation settings (monolingual, translated, zero-shot cross-lingual). Despite this breadth, MEGA reports significant performance gaps between Turkish and English, as well as among other non-English languages.
Recent efforts such as M2QA \citep{englander2024m2qa} and SeaEval \citep{wang2023seaeval} show that LLM performance can vary significantly by domain and language. Additionally, large-scale multilingual classification benchmarks have been developed \citep{ma-etal-2025-taxi1500,adelani-etal-2024-sib}. Notably, TUMLU \citep{isbarov2025tumlu} evaluates LLMs across eight Turkic languages and 11 school-subject domains, offering a culturally aware assessment framework for Turkic language models.

\subsection{Benchmarks tailored for Turkish}
Several benchmarks have been designed specifically for Turkish NLP. Mukayese \citep{safaya-etal-2022-mukayese} includes seven tasks, primarily targeting the evaluation of pre-LLM-era multilingual models using fine-tuning. Benchmarks such as \citet{acikgoz-etal-2024-bridging} and \citet{openllm-Turkish-leaderboard} are built from machine-translated datasets, including ARC \citep{clark2018thinksolvedquestionanswering}, TruthfulQA \citep{lin-etal-2022-truthfulqa}, and GSM8K \citep{cobbe2021gsm8k}. TurkishMMLU \citep{yuksel2024turkishmmlu}, a localized version of MMLU, provides 10K high-school-level questions spanning nine subjects with zero-shot and few-shot evaluations. TR-MMLU \citep{bayram2024setting} expands this further with 6,200 questions across 62 categories, including law and healthcare.

\cetvel~advances beyond these resources in three key ways:
\begin{enumerate}[label=(\roman*),nosep]
\item it covers a broader set of 23 tasks that include both discriminative and generative settings, unlike MCQA-heavy benchmarks such as TurkishMMLU, TR-MMLU, and TUMLU;
\item it includes tasks explicitly designed around Turkish linguistic and cultural content, a critical shortcoming of machine-translated benchmarks;
\item it is more comprehensive and up-to-date than Mukayese, supporting zero-shot evaluation of modern LLMs.
\end{enumerate}

%% file: sections/sec_datasets.tex
\section{Tasks and Datasets}
\label{sec:datasets}
\cetvel~includes a diverse set of tasks designed to comprehensively evaluate large language models in Turkish. These tasks are grouped into two high-level categories: natural language understanding (NLU) and natural language generation (NLG). In total, \cetvel~spans 23 tasks drawn from publicly available benchmarks and curated datasets, with particular emphasis on linguistic and cultural relevance.
Figure~\ref{fig:data_dist} provides a visual breakdown of task categories and subtypes.

\subsection{Language Understanding Tasks}
We organize NLU tasks into four subcategories: (i) extractive question answering (QA), (ii) multiple-choice question answering (MCQA), (iii) text classification (TC), and (iv) natural language inference (NLI).

\subsubsection*{Extractive Question Answering}
In extractive QA, the model is presented with a question and a contextual passage that contains the answer. The objective is to extract the correct answer span from the context. \cetvel~includes the following resources: \textbf{XQuAD} \citep{artetxe-etal-2020-cross}  \citep{artetxe-etal-2020-cross} extends the English SQuAD dataset \citep{rajpurkar-etal-2016-squad} with crowd-sourced translations into 11 languages, including Turkish. \textbf{MKQA} \citep{longpre-etal-2021-mkqa} offers 10K aligned question–answer pairs across 26 languages. It lacks contextual passages, we retain this original format. \textbf{TQuAD} contains context-based questions on Turkish and Islamic history, making it uniquely suited for culturally grounded QA in Turkish\footnote{\fngithub{https://github.com/TQuad/turkish-nlp-qa-dataset}}.

\subsubsection*{Multiple Choice Question Answering}
The multiple‐choice question answering (MCQA) is well-suited for zero-shot evaluation, and hence it serves as the main format in many recent NLP benchmarks. In \cetvel, we include datasets spanning three subdomains:

\begin{enumerate}[label=(\roman*),nosep]
\item \textbf{Exam-style Assessments:} \textbf{Exams} \citep{hardalov2020exams} features 393 questions drawn from Turkish high-school subjects such as mathematics and religion. \textbf{TurkishMMLU} \citep{yuksel2024turkishmmlu} is an adaptation of MMLU offering 900 questions across a wide range of academic domains. \textbf{Belebele} \citep{bandarkar-etal-2024-belebele} includes 900 reading comprehension questions, translated by professionals into 122 languages, including Turkish.
\item \textbf{Procedural and Commonsense Reasoning:} \textbf{Turkish-PLU} \citep{uzunoglu-sahin-2023-benchmarking} contains four tasks adapted from WikiHow, including goal inference, next-event prediction, step inference, and step ordering. \textbf{XCOPA} \citep{ponti2020xcopa} is a multilingual benchmark requiring causal reasoning, in which models must infer either a cause or an effect given a premise.
\item \textbf{Specific to Turkish:} \textbf{Turkish Proverbs}\footnote{\huggingface{https://huggingface.co/datasets/furkanunluturk/turkce-atasozleri}} comprises 1,730 Turkish proverbs paired with definitions from official linguistic resources. Distractors are generated using Llama3.3-70B embeddings. 
\textbf{BilmeceBench}\footnote{\huggingface{https://huggingface.co/datasets/selimc/bilmecebench}} has 442 riddles converted into MCQA format with randomized distractors. \textbf{CircumflexTR} is curated specifically for \cetvel, and targets minimal pairs distinguished by the circumflex diacritic (e.g., kar ``snow'' vs. kâr ``interest'').
\end{enumerate}

\subsubsection*{Text Classification}
We frame text classification tasks as MCQA by presenting labels as choices. The task involves selecting the most appropriate label for a given input. We include the following five datasets: %
\textbf{OffensEval} \citep{coltekin-2020-corpus} for hate speech detection, performed on user-generated social media data. \textbf{IronyTR} \citep{ozturk2021ironytr} for detecting irony within sentences. \textbf{STSb-TR}, a machine-translated variant of the STS benchmark, for predicting semantic similarity between two given languages.

\subsubsection*{Natural Language Inference}
Natural language inference (NLI) involves predicting the logical relationship (entailment, contradiction, or neutrality) between a premise and a hypothesis. Similar to the text classification tasks, we treat the NLI task as a multiple-choice question answering task where we use Turkish versions of the widely adopted \textbf{XNLI} \citep{conneau2018xnli}, \textbf{SNLI} \citep{bowman-etal-2015-large}, and \textbf{MNLI} \citep{budur-etal-2020-data} datasets.

\subsection{Language Generation Tasks}
We include three generation tasks: summarization, machine translation, and grammatical error correction.

\subsubsection*{Summarization}
The goal is to generate a concise summary from a paragraph-length input. We evaluate models on the Turkish portions of \textbf{MLSum} \citep{scialom-etal-2020-mlsum} (news summaries), \textbf{XLSum} \citep{Hasan_2021} (single-sentence summaries from BBC articles), and \textbf{WikiLingua} \citep{ladhak-etal-2020-wikilingua} (step-by-step instructional summaries from WikiHow).

\input{tables/results1_py_top10}

\subsubsection*{Machine Translation}
We also evaluate English-to-Turkish translation using \textbf{WMT-16} \citep{bojar-EtAl:2016:WMT1}. This task measures cross-lingual language processing and Turkish generation quality.

\subsubsection*{Grammatical Error Correction}
This task involves correcting grammatical mistakes in Turkish sentences. We use \textbf{GECTurk} \citep{kara-etal-2023-gecturk}, which contains 22k sentence pairs synthetically generated using 25 expert-defined grammar rules.

%% file: tables/results1_py_top10.tex
\renewcommand{\arraystretch}{1.15}{
\begin{table*}[!t]
\centering
\begin{tabular}{lrrrrrrrr}
\toprule
 & \textbf{QA} & \textbf{MC} & \textbf{TC} & \textbf{NLI} & \textbf{SUM} & \textbf{MT} & \textbf{GEC} & \textbf{Avg.} \\
\hline
\bluesquare\hspace{2mm}Llama-3.3-70B-Instruct & 16.1 & \textbf{60.1} & \textbf{58.1} & 32.4 & 16.2 & 13.6 & 44.1 & \textbf{34.4} \\
\orangesquare\hspace{2mm}Aya-Expanse-32B & 26.2 & 55.6 & 55.3 & \textbf{43.3} & \textbf{22.4} & \textbf{20.1} & 4.5 & 32.5 \\
\orangesquare\hspace{2mm}Aya-23-35B & 23.7 & 48.8 & 38.0 & 37.6 & 17.6 & 18.5 & 30.8 & 30.7 \\
\bluedot\hspace{2mm}Llama-3.1-8B & 19.3 & 45.8 & 44.8 & 32.2 & 13.5 & 15.6 & 35.3 & 29.5 \\
\bluesquare\hspace{2mm}Llama-3.1-8B-Instruct & 18.0 & 50.1 & 40.1 & 36.0 & 13.5 & 15.6 & 31.5 & 29.3 \\
\orangedot\hspace{2mm}Qwen2.5-7B & 20.5 & 50.6 & 51.6 & 34.0 & 12.8 & 5.5 & 22.3 & 28.2 \\
\bluedot\hspace{2mm}Llama-3-8B & 20.9 & 43.0 & 40.6 & 33.9 & 12.3 & 11.3 & 34.1 & 28.0 \\
\redsquare\hspace{2mm}Cere-Llama-3-8B & 24.2 & 44.8 & 43.7 & 34.0 & 3.5 & 0.1 & \textbf{46.0} & 28.0 \\
\bluesquare\hspace{2mm}Ministral-2410-8B-Instruct & 14.2 & 42.8 & 38.0 & 34.0 & 12.8 & 11.2 & 39.1 & 27.5 \\
\orangedot\hspace{2mm}Qwen2.5-14B & \textbf{26.7} & 52.6 & 37.7 & 34.0 & 13.0 & 8.1 & 18.9 & 27.3 \\
\bottomrule
\end{tabular}
\caption{Performances of the top-10 models. Bold-face indicates best performances. Shapes next to model ids denote the model type: Base pretrained models are represented by circles and instruction-tuned LLMs are denoted by squares. Colors indicate the language focus: \textcolor{blue}{blue} for English-centric, \textcolor{orange}{yellow} for multilingually-pretrained, and \textcolor{red}{red} for Turkish-centric LLMs. \textbf{QA} denotes extractive QA, \textbf{MC} denotes multiple-choice QA, \textbf{TC} denotes text classification, \textbf{NLI} denotes natural language inference, \textbf{SUM} denotes summarization, \textbf{MT} denotes machine translation, and \textbf{GEC} denotes grammatical error correction.  Llama-3.3-70B-Instruct achieves the best overall performance, followed by Aya-Expanse-32B. Among Turkish-centric LLMs, Cere-Llama-3-8B achieves an exceptional performance, even surpassing Llama-3.3-70B-Instruct on GEC and extractive QA about Turkish history.}
\label{tab:top10}
\end{table*}
}

%% file: sections/sec_experimental_setup.tex
\section{Experimental Setup}
\label{sec:experimental_setup}

This section outlines the models evaluated in \cetvel, the metrics used for assessment, and implementation details of our experimental pipeline.

\subsection{Models}
We evaluate 33 open-weight models collected from the Huggingface Transformers package \citep{wolf-etal-2020-transformers}. Models are grouped into three main categories based on language coverage and pretraining objectives:

\subsubsection*{General-purpose LLMs}
These models are primarily pretrained on English but might include additional language data during pretraining.
For this category of models, we cover \textbf{Mistral} \citep{jiang2023mistral7b}, \textbf{Mixtral} \citep{jiang2024mixtral} and \textbf{Llama 3} \citep{dubey2024llama3herdmodels}.

\subsubsection*{Multilingual LLMs}
These models are pretrained to support a wide range of languages. We include \textbf{Aya-101} \citep{ustun-etal-2024-aya}, \textbf{Aya-23} \citep{aryabumi2024aya}, \textbf{Aya-Expanse} \citep{dang2024aya}, \textbf{Llama 3.1}, \textbf{Llama 3.2}, \textbf{Llama 3.3} \citep{dubey2024llama3herdmodels}, and \textbf{Qwen2.5} \citep{yang2024qwen2} models.

\subsubsection*{Turkish-centric LLMs}
These models are either pretrained exclusively on Turkish or further finetuned on Turkish data.
We include \textbf{Kanarya} \citep{safaya-etal-2022-mukayese}, \textbf{Turna} \citep{uludogan-etal-2024-turna}, \textbf{Commencis-LLM-7B} \citep{Commencis-2024}, \textbf{Trendyol-LLM-7B} \citep{Trendyol-2024}, and \textbf{Cere-Llama-3-8B} \citep{CerebrumTech-2024}.
Kanarya and Turna models are pretrained on solely Turkish.
The remaining three models, Commencis-LLM-7B, Trendyol-LLM-7B and Cere-Llama-3-8B are finetuned on Turkish instruction following data by Turkish tech companies.
Specifically, Commencis-LLM-7B and Trendyol-LLM-7B use Mistral-7B as base model, and Cere-Llama-3-8B is built upon Llama3-8B model.

We further categorize models by their architecture (decoder-only vs. encoder-decoder), training paradigm (pretraining-only vs. instruction-tuned), and parameter count.
Within \cetvel, all of the evaluated LLMs have fewer than 70B parameters and are open-weights, publicly available models.
Exceptionally, Turna and Aya-101 models employ an encoder-decoder architecture built upon T5 \citep{raffel2020exploring}.

\subsection{Evaluation Metrics}
We use standard automatic metrics tailored to each task type. \textbf{Language understanding tasks} are evaluated using \textbf{accuracy}. For MCQA tasks, candidate answers are scored based on per-token perplexity, and the option with the lowest perplexity is selected. \textbf{Extractive QA} is evaluated using \textbf{Exact Match (EM)} \citep{rajpurkar-etal-2016-squad}. \textbf{Summarization} is evaluated using ROUGE-2 \citep{lin-2004-rouge}. \textbf{Machine translation} using \textbf{BLEU-4} \citep{papineni-etal-2002-bleu}, and
\textbf{Grammatical Error Correction} using \textbf{macro-F1}. We do not employ LLM-as-a-judge metrics \citep{zheng2023judging}, as they have been shown to be unreliable in multilingual settings \citep{fu2025reliablemultilingualllmasajudge}.

\subsection{Implementation Details}
All experiments are conducted using the \textbf{LM Evaluation Harness} \citep{eval-harness}, a framework that supports evaluation of Huggingface-compatible models and integrates with the \textbf{vLLM} inference backend \citep{kwon2023efficient} for efficient model serving.
For NLU tasks, we use a batch size of 4. For generation tasks, we process one instance at a time and limit outputs to a maximum of 64 tokens, following the protocol used in Mukayese \citep{safaya-etal-2022-mukayese}.
We use \textbf{beam search} decoding with a beam width of 5 across all generative tasks, ensuring deterministic evaluation.
We run experiments for each single model on eight NVIDIA A40 GPUs. Experiment duration depends on the model size, for instance, the entire set of experiments for an 8 billion parameter model completes less than two days.
We note that, we conduct each experiment with exactly one single forward run per model.

\begin{figure*}[!t]
    \includegraphics[width=1.0\textwidth]{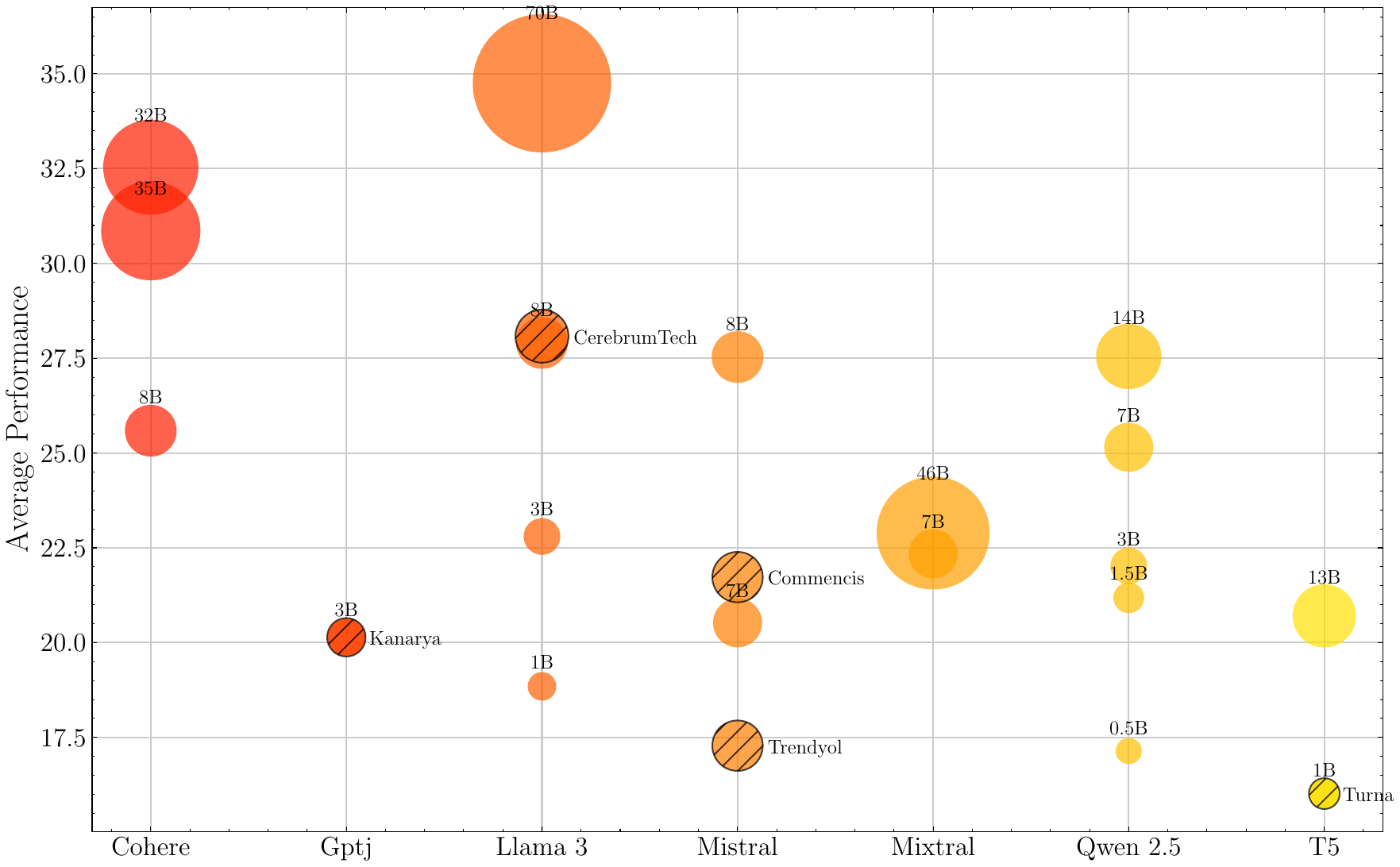}
    \caption{Overall performances on \cetvel~grouped by model family. Model size is indicated by the size of the corresponding sphere. A striped sphere indicates that the corresponding model is Turkish-centric LLM. Our experiments reveal that Llama-3.3-70B-Instruct achieves the best overall performance }
    \label{fig:model_performances}
\end{figure*}

%% file: sections/sec_results.tex
\section{Results}
\label{sec:results}

\begin{figure*}[t]
    \includegraphics[width=1.0\textwidth]{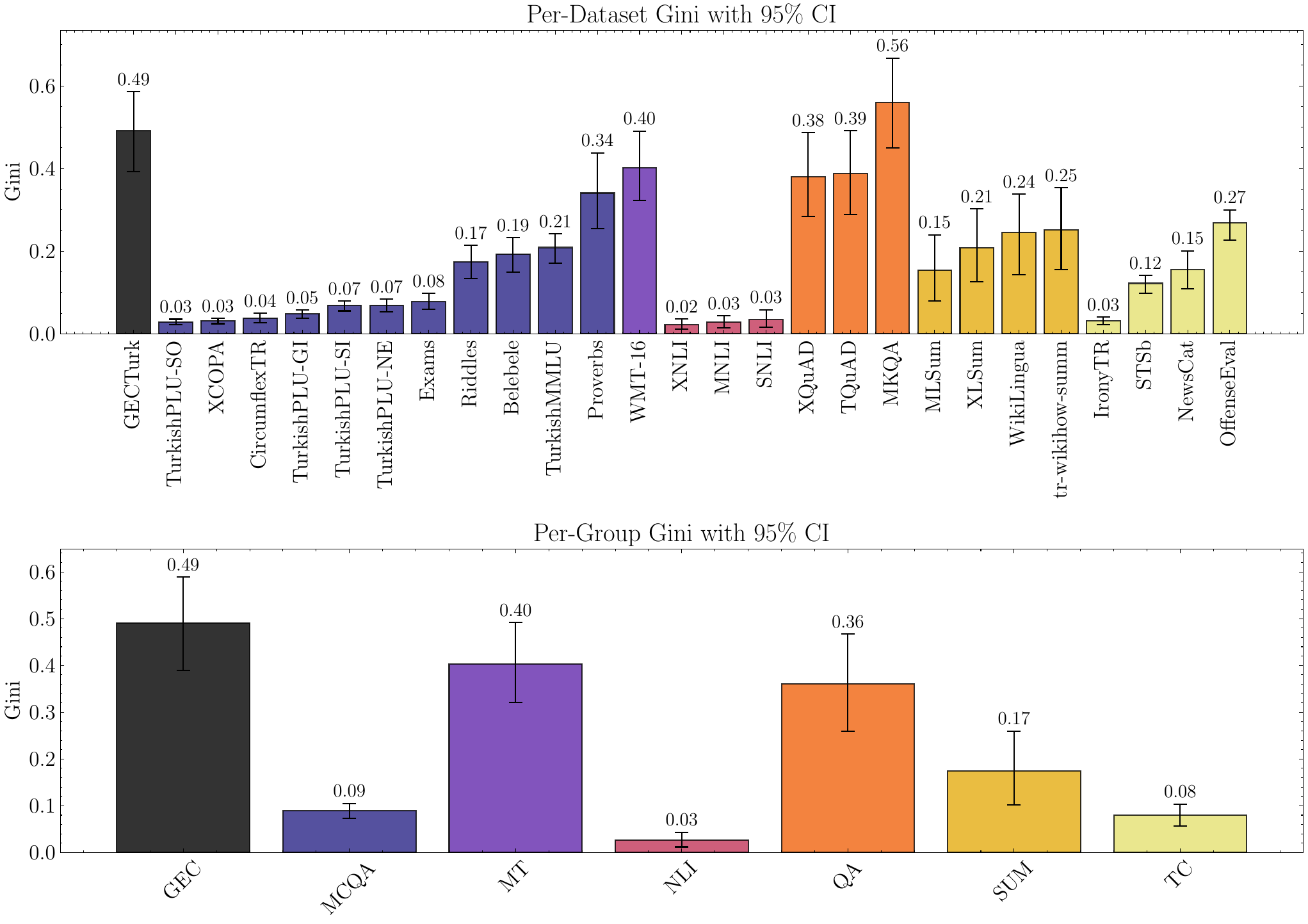}
    \caption{Bar plots with 95\% CI of the task-specific and category-wise bootstrapped Gini coefficients. We find that grammatical error correction, machine translation and extractive question answering tasks are the strongest indicators of differentiating model performances. Conversely, NLI and TC tasks contribute the least to differentiating model performances.}
    \label{fig:gini}
\end{figure*}

We present our evaluation results, focusing on model performance with respect to parameter size, multilingual coverage, and training paradigm. Table~\ref{tab:top10} shows the top-20 models\footnote{Full category-specific results are provided in the appendix.} across all task categories, and Figure~\ref{fig:model_performances} visualizes average performance grouped by model architecture and size.

\subsection{Overall Results}
Our results indicate that LLaMA 3 models consistently outperform alternatives within comparable parameter ranges. The best-performing model overall is \textbf{Llama-3.3-70B-Instruct}, which exceeds the second-best model, \textbf{Aya-Expanse-32B}, by 4.5 points in average score. Notably, \textbf{Llama-3.1-8B} performs comparably to larger models such as \textbf{Aya-23-35B} and \textbf{Aya-Expanse-32B}, indicating strong performance scaling efficiency. Unexpectedly, base pretrained \textbf{Qwen2.5} models outperform their instruction-tuned variants across all parameter sizes, except for the smallest 0.5B model. 

Turkish-centric instruction-tuned models generally lag behind multilingual and English-centric models. In particular, \textbf{Commencis-LLM-7B}~and \textbf{Trendyol-LLM-7B} underperform relative to their base model \textbf{Mistral-7B}. Models pretrained from scratch in Turkish also show weak results: \textbf{Turna-1B} ranks last overall, and \textbf{Kanarya-2B} achieves only an average score of 19.9. One exception~among~Turkish-centric models is \textbf{Cere-Llama-3-8B}, which excels in \textbf{grammatical error correction} and \textbf{extractive QA} on culturally specific datasets (e.g., TQuAD), outperforming even Llama-3.3-70B-Instruct on these tasks. However, Cere-Llama-3-8B underperforms in \textbf{machine translation} and \textbf{knowledge-intensive tasks}, likely due to the lack of English exposure and general-domain fine-tuning. This highlights the importance of including cross-lingual and domain-diverse data during instruction tuning for low-resource language models.

Overall, \cetvel~reveals a substantial gap in the current instruction-tuning strategies for Turkish and highlights the limitations of monolingual or narrowly focused Turkish-centric models.

\subsection{Language Understanding Tasks}
For non-generative tasks, \textbf{Llama-3.3-70B-Instruct} again leads overall, particularly on knowledge-intensive benchmarks such as Turkish proverbs and riddles.
Larger models tend to outperform smaller ones on exam-style tasks (TurkishMMLU, Belebele, Exams), likely due to increased memorization and reasoning capacity. 
However, performance on commonsense reasoning (e.g., XCOPA) is less sensitive to model size.
For instance, \textbf{Qwen2.5-0.5B} achieves 53.6\% accuracy on XCOPA, just 14.4 points below the strongest model, \textbf{Qwen2.5-14B}.
On the extractive QA task, \textbf{Qwen2.5} models outperform LLaMA models on XQuAD, with \textbf{Qwen2.5-14B} ranking highest.
For TQuAD, however, \textbf{Cere-Llama-3-8B} achieves the best score, outperforming all others despite its smaller size—highlighting the benefits of task-specific tuning for culturally grounded datasets.

\subsection{Language Generation Tasks}
On generative tasks, \textbf{Aya} models (excluding \textbf{Aya-101-8B}) take the lead in summarization and machine translation, likely due to their strong multilingual pretraining.
These models particularly benefit from overlapping multilingual content in training corpora (e.g., WikiLingua), which may enhance memorization and transfer.
In contrast, \textbf{Cere-Llama-3-8B} achieves the best results on grammatical error correction even surpassing 70b parameters model Llama-3.3-70B-Instruct, but performs poorly in summarization and translation tasks-again pointing to the importance of balanced cross-lingual training.
All remaining Turkish-centric models perform extremely poorly on machine translation due to exclusion of English during instruction-tuning phase, where Kanarya-2B is the highest performing model, attaining a BLUE-4 score of $3.1$.
Nonetheless, Kanarya-2B achieves an overall performance of 24.7 ROGUE-2 score on MLSum, outperforming same parameter-scale Qwen2.5 models.

\subsection{Turkish-Centric LLMs}
Overall, Turkish-centric models fall behind English-centric and multilingual LLMs.
This is the case for both LLMs pretrained from scratch on Turkish or LLMs instruction-tuned on Turkish after pretraining.
Turkish LLMs pretrained from scratch, Turna-1B and Kanarya-2B, rank in the lower places.
LLMs finetuned on Turkish instructions perform better, yet they underperform against their base models.
Both Trendyol-LLM-7B and Commencis-LLM-7B models achieve overall performances below their base LLM, Mistral-7B.
As we mentioned earlier, they perform extremely poorly on machine translation, due to catastrophic forgetting English \citep{liu-etal-2024-catastrophic}.
These models attain mediocre overall performances, highlighting that there is a large room for improvement in developing LLMs that can effectively process Turkish.
As we mentioned earlier, Cere-Llama-3-8B achieves the highest score on TQuAD and GECturk datasets, even outperforming the largest model Llama-3.3-70B-Instruct.
Nonetheless, Cere-Llama-3-8B also suffers from catastrophic forgetting like other LLMs instruction-tuned on Turkish.

\subsection{Task Discrimination Analysis}
To assess which tasks most effectively differentiate model capabilities, we compute bootstrapped Gini coefficients \citep{dorfman1979formula} across task categories and within the tasks using the following formula, 

\begin{equation}\label{eq:gini}
G \;=\; \frac{1}{2\,n^{2}\,\bar x}
\sum_{i=1}^{n}\sum_{j=1}^{n} \bigl|x_i - x_j\bigr|,
\quad
\bar x = \frac{1}{n}\sum_{k=1}^n x_k.
\end{equation}

As shown in Figure \ref{fig:gini}, Grammatical error correction (\textbf{GEC}), machine translation (\textbf{MT}), and question answering (\textbf{QA}) exhibit the highest discrimination power, with coefficients of 0.490, 0.402, and 0.362, respectively. These tasks consistently produce wide performance gaps across models, making them particularly informative for benchmarking. Conversely, natural language inference (\textbf{NLI}), text classification (\textbf{TC}), and multiple-choice question answering (\textbf{MCQA}) have much lower Gini coefficients (0.039, 0.080, and 0.090, respectively), suggesting limited utility in differentiating LLMs in the Turkish setting.

Additionally, some task categories contain tasks with substantially varying discriminative power. For example, the \textbf{XCOPA} and \textbf{Proverbs} tasks, both from the \textbf{MCQA} category, have Gini coefficients of 0.03 and 0.34, respectively, indicating room for improvement in this category. Despite this variability, tasks from the \textbf{QA} and \textbf{NLI} categories consistently show high and low Gini coefficients, respectively, which may indicate overall category-level discrimination power rather than dataset selection.

%% file: sections/sec_conclusion.tex
\section{Conclusion}
\label{sec:conclusion}

In this work, we introduced \cetvel, a task-diverse NLP benchmark designed to evaluate large language models in Turkish, with particular attention to linguistic and cultural specificity. \cetvel~addresses key limitations of previous efforts, which often lacked task diversity or overlooked culturally grounded content.

By incorporating underexplored phenomena such as proverbs and riddles, \cetvel~broadens the scope of evaluation beyond standard NLP tasks and provides a more comprehensive testbed for both multilingual and Turkish-specific models.

Our extensive experiments reveal that instruction-tuned Turkish LLMs consistently underperform compared to general-purpose models that have been pretrained on multilingual corpora including Turkish. These results point to the need for more effective instruction-tuning strategies tailored to Turkish, including higher-quality prompts, culturally relevant tasks, and improved validation pipelines.

Among the evaluated models, Llama 3 variants deliver the strongest overall performance across tasks and parameter ranges. Furthermore, our task discrimination analysis shows that Grammar Error Correction, Machine Translation, and Question Answering are particularly effective in distinguishing model capabilities, while NLI and Topic Classification tasks contribute less to differentiation.

We hope \cetvel~serves as a valuable resource for advancing Turkish NLP and guiding the development of more robust and culturally aware LLMs.

\subsection*{Limitations}
In this version of \cetvel, our evaluation is limited to open-weight models with up to 70B parameters. Due to recent API restrictions, we were unable to include widely used proprietary models that no longer support log-probability outputs, which are essential for our evaluation pipeline.
Moreover, our analysis is restricted to the zero-shot setting. While this provides a controlled and reproducible baseline, incorporating one-shot and few-shot evaluations remains an important direction for future iterations of the benchmark.
Finally, we note that \cetvel~might include user-generated web data, which can be noisy as recently shown by \citet{cengiz2025evaluatingqualitybenchmarkdatasets}.
Nevertheless, we retain these data resources because the underlying information for solving the tasks remains accurate.

\subsection*{Acknowledgments}
We thank Mustafa Cemil Güney and Demir Ekin Arıkan for their contributions on the early stages of the development.
Ilker Kesen was supported by the European Union’s Horizon 2020 research and innovation program under grant agreement No. 101135671 (TrustLLM). 
This work has been supported by the Scientific and Technological Research Council of
Türkiye (TÜBITAK) as part of the project “Automatic Learning of Procedural Language from Natural Language Instructions for Intelligent Assistance” with the number 121C132.
We also gratefully acknowledge KUIS AI Center for providing computational support.

%% file: tables/results1_py.tex
\renewcommand{\arraystretch}{1.05}{
\begin{table*}[ht]
\centering
\begin{tabular}{lrrrrrrrr}
\toprule
\textbf{} & \textbf{QA} & \textbf{MC} & \textbf{TC} & \textbf{NLI} & \textbf{SUM} & \textbf{MT} & \textbf{GEC} & \textbf{Avg.} \\
\hline
\bluesquare\hspace{2mm}Llama-3.3-70B-Instruct & 16.1 & \textbf{60.1} & \textbf{58.1} & 32.4 & 16.2 & 13.6 & 44.1 & \textbf{34.4} \\
\orangesquare\hspace{2mm}Aya-Expanse-32B & 26.2 & 55.6 & 55.3 & \textbf{43.3} & \textbf{22.4} & \textbf{20.1} & 4.5 & 32.5 \\
\orangesquare\hspace{2mm}Aya-23-35B & 23.7 & 48.8 & 38.0 & 37.6 & 17.6 & 18.5 & 30.8 & 30.7 \\
\bluedot\hspace{2mm}Llama-3.1-8B & 19.3 & 45.8 & 44.8 & 32.2 & 13.5 & 15.6 & 35.3 & 29.5 \\
\bluesquare\hspace{2mm}Llama-3.1-8B-Instruct & 18.0 & 50.1 & 40.1 & 36.0 & 13.5 & 15.6 & 31.5 & 29.3 \\
\orangedot\hspace{2mm}Qwen2.5-7B & 20.5 & 50.6 & 51.6 & 34.0 & 12.8 & 5.5 & 22.3 & 28.2 \\
\bluedot\hspace{2mm}Llama-3-8B & 20.9 & 43.0 & 40.6 & 33.9 & 12.3 & 11.3 & 34.1 & 28.0 \\
\redsquare\hspace{2mm}Cere-Llama-3-8B & 24.2 & 44.8 & 43.7 & 34.0 & 3.5 & 0.1 & \textbf{46.0} & 28.0 \\
\bluesquare\hspace{2mm}Ministral-2410-8B-Instruct & 14.2 & 42.8 & 38.0 & 34.0 & 12.8 & 11.2 & 39.1 & 27.5 \\
\orangedot\hspace{2mm}Qwen2.5-14B & \textbf{26.7} & 52.6 & 37.7 & 34.0 & 13.0 & 8.1 & 18.9 & 27.3 \\
\orangesquare\hspace{2mm}Aya-Expanse-8B & 15.1 & 52.2 & 43.6 & 34.7 & 12.2 & 17.0 & 5.5 & 25.7 \\
\orangesquare\hspace{2mm}Aya-23-8B & 18.5 & 45.7 & 45.3 & 33.3 & 12.8 & 16.9 & 4.3 & 25.2 \\
\bluedot\hspace{2mm}Llama-3.2-3B & 14.2 & 40.6 & 38.7 & 32.5 & 12.1 & 7.9 & 26.8 & 24.7 \\
\orangesquare\hspace{2mm}Qwen2.5-14B-Instruct & 0.4 & 56.0 & 43.3 & 35.2 & 12.8 & 12.4 & 10.9 & 24.4 \\
\bluesquare\hspace{2mm}Llama-3-8B-Instruct & 9.0 & 49.3 & 40.6 & 33.8 & 13.8 & 13.6 & 10.0 & 24.3 \\
\orangedot\hspace{2mm}Qwen2.5-3B & 19.7 & 41.5 & 40.2 & 34.0 & 12.3 & 4.1 & 14.7 & 23.8 \\
\bluedot\hspace{2mm}Mistral-v0.3-7B & 16.8 & 39.0 & 37.7 & 34.0 & 7.9 & 3.8 & 23.5 & 23.2 \\
\orangedot\hspace{2mm}Qwen2.5-5B & 22.1 & 39.8 & 35.3 & 33.9 & 12.0 & 4.1 & 14.2 & 23.0 \\
\bluesquare\hspace{2mm}Mixtral-v0.1-7B-Instruct & 11.0 & 45.1 & 45.6 & 33.7 & 10.9 & 9.5 & 3.6 & 22.8 \\
\orangesquare\hspace{2mm}Qwen2.5-7B-Instruct & 0.5 & 50.9 & 48.4 & 35.0 & 11.0 & 5.4 & 6.6 & 22.5 \\
\bluedot\hspace{2mm}Mistral-v0.1-7B & 16.6 & 39.2 & 37.3 & 33.9 & 3.6 & 5.0 & 20.8 & 22.4 \\
\bluesquare\hspace{2mm}Llama-3.2-3B-Instruct & 15.5 & 39.1 & 36.9 & 34.0 & 8.1 & 4.9 & 16.7 & 22.2 \\
\reddot\hspace{2mm}Commencis-7B & 5.1 & 38.1 & 39.4 & 32.5 & 9.8 & 4.7 & 23.7 & 21.9 \\
\orangesquare\hspace{2mm}Aya-101-13B & 5.2 & 40.6 & 42.3 & 27.8 & 0.1 & 0.0 & 32.4 & 21.2 \\
\bluedot\hspace{2mm}Llama-3.2-1B & 4.0 & 37.7 & 43.7 & 34.1 & 7.5 & 1.2 & 18.1 & 20.9 \\
\redsquare\hspace{2mm}Kanarya-2B & 1.0 & 41.0 & 47.8 & 33.1 & 10.9 & 3.0 & 6.6 & 20.5 \\
\orangesquare\hspace{2mm}Qwen2.5-3B-Instruct & 3.1 & 45.2 & 35.7 & 34.0 & 12.0 & 6.6 & 4.1 & 20.1 \\
\orangesquare\hspace{2mm}Qwen2.5-5B-Instruct & 10.4 & 42.8 & 33.0 & 33.9 & 12.0 & 4.7 & 3.6 & 20.1 \\
\bluesquare\hspace{2mm}Mistral-v0.3-7B-Instruct & 8.7 & 38.6 & 42.5 & 31.7 & 11.2 & 5.9 & 1.0 & 19.9 \\
\orangesquare\hspace{2mm}Qwen2.5-5B-Instruct & 10.1 & 36.9 & 27.8 & 33.9 & 10.7 & 3.1 & 1.8 & 17.7 \\
\reddot\hspace{2mm}Trendyol-v1.0-7B-Base & 0.3 & 38.4 & 41.8 & 34.3 & 5.3 & 1.0 & 0.1 & 17.3 \\
\orangedot\hspace{2mm}Qwen2.5-5B & 3.8 & 37.8 & 30.7 & 33.9 & 11.3 & 1.4 & 0.9 & 17.1 \\
\reddot\hspace{2mm}Turna-1B & 0.0 & 35.9 & 36.6 & 34.1 & 7.1 & 0.1 & 0.0 & 16.3 \\
\hline
\end{tabular}
\caption{Overall results of the models, sorted by their average scores. Base and instruction-tuned variants are represented by circles and squares, respectively. Colors indicate the language focus: blue for English-focused, yellow for multilingual-focused, and red for Turkish-focused models.}
\label{appendix_overall}
\end{table*}
}

%% file: tables/gec.tex
\renewcommand{\arraystretch}{1.05}{
\begin{table*}[ht]
\centering
\begin{tabular}{lrr}
\toprule
\textbf{} & \textbf{GECTurk} & \textbf{Avg.} \\
\hline
\redsquare\hspace{2mm}Cere-Llama-3-8B & \textbf{46.0} & \textbf{46.0} \\
\bluesquare\hspace{2mm}Llama-3.3-70B-Instruct & 44.1 & 44.1 \\
\bluesquare\hspace{2mm}Ministral-2410-8B-Instruct & 39.1 & 39.1 \\
\bluedot\hspace{2mm}Llama-3.1-8B & 35.3 & 35.3 \\
\bluedot\hspace{2mm}Llama-3-8B & 34.1 & 34.1 \\
\orangesquare\hspace{2mm}Aya-101-13B & 32.4 & 32.4 \\
\bluesquare\hspace{2mm}Llama-3.1-8B-Instruct & 31.5 & 31.5 \\
\orangesquare\hspace{2mm}Aya-23-35B & 30.8 & 30.8 \\
\bluedot\hspace{2mm}Llama-3.2-3B & 26.8 & 26.8 \\
\reddot\hspace{2mm}Commencis-7B & 23.7 & 23.7 \\
\bluedot\hspace{2mm}Mistral-v0.3-7B & 23.5 & 23.5 \\
\orangedot\hspace{2mm}Qwen2.5-7B & 22.3 & 22.3 \\
\bluedot\hspace{2mm}Mistral-v0.1-7B & 20.8 & 20.8 \\
\orangedot\hspace{2mm}Qwen2.5-14B & 18.9 & 18.9 \\
\bluedot\hspace{2mm}Llama-3.2-1B & 18.1 & 18.1 \\
\bluesquare\hspace{2mm}Llama-3.2-3B-Instruct & 16.7 & 16.7 \\
\orangedot\hspace{2mm}Qwen2.5-3B & 14.7 & 14.7 \\
\orangedot\hspace{2mm}Qwen2.5-5B & 14.2 & 14.2 \\
\orangesquare\hspace{2mm}Qwen2.5-14B-Instruct & 10.9 & 10.9 \\
\bluesquare\hspace{2mm}Llama-3-8B-Instruct & 10.0 & 10.0 \\
\redsquare\hspace{2mm}Kanarya-2B & 6.6 & 6.6 \\
\orangesquare\hspace{2mm}Qwen2.5-7B-Instruct & 6.6 & 6.6 \\
\orangesquare\hspace{2mm}Aya-Expanse-8B & 5.5 & 5.5 \\
\orangesquare\hspace{2mm}Aya-Expanse-32B & 4.5 & 4.5 \\
\orangesquare\hspace{2mm}Aya-23-8B & 4.3 & 4.3 \\
\orangesquare\hspace{2mm}Qwen2.5-3B-Instruct & 4.1 & 4.1 \\
\bluesquare\hspace{2mm}Mixtral-v0.1-7B-Instruct & 3.6 & 3.6 \\
\orangesquare\hspace{2mm}Qwen2.5-5B-Instruct & 3.6 & 3.6 \\
\orangesquare\hspace{2mm}Qwen2.5-5B-Instruct & 1.8 & 1.8 \\
\bluesquare\hspace{2mm}Mistral-v0.3-7B-Instruct & 1.0 & 1.0 \\
\orangedot\hspace{2mm}Qwen2.5-5B & 0.9 & 0.9 \\
\reddot\hspace{2mm}Trendyol-v1.0-7B-Base & 0.1 & 0.1 \\
\reddot\hspace{2mm}Turna-1B & 0.0 & 0.0 \\
\hline
\end{tabular}
\caption{Grammatical error correction results of the models, sorted by their average scores. Base and instruction-tuned variants are represented by circles and squares, respectively. Colors indicate the language focus: blue for English-focused, yellow for multilingual-focused, and red for Turkish-focused models.}
\label{appendix_gec}
\end{table*}
}

%% file: tables/mcqa.tex
\renewcommand{\arraystretch}{1.05}{
\begin{table*}[ht]
\centering
\resizebox{\textwidth}{!}{%
\begin{tabular}{lrrrrrrrrrrrrr}
\toprule
\textbf{} & \textbf{XCOPA} & \textbf{PLU} & \textbf{PLU\_GI} & \textbf{PLU\_NEP} & \textbf{PLU\_SI} & \textbf{PLU\_SO} & \textbf{Exams} & \textbf{Belebele} & \textbf{Proverbs} & \textbf{TurkishMMLU} & \textbf{BilmeceBench} & \textbf{CircumflexTR} & \textbf{Avg.} \\
\hline
\bluesquare\hspace{2mm}Llama-3.3-70B-Instruct & 65.0 & \textbf{54.2} & \textbf{49.1} & \textbf{58.0} & 39.1 & 65.1 & \textbf{39.2} & \textbf{86.8} & \textbf{92.5} & \textbf{64.6} & \textbf{72.6} & \textbf{67.1} & \textbf{60.1} \\
\orangesquare\hspace{2mm}Qwen2.5-14B-Instruct & \textbf{66.6} & 48.5 & 40.6 & 49.8 & 35.1 & 62.2 & 29.8 & 84.7 & 78.3 & 59.4 & 57.0 & 58.6 & 56.0 \\
\orangesquare\hspace{2mm}Aya-Expanse-32B & 59.2 & 51.8 & 44.8 & 55.1 & 39.2 & 63.0 & 36.9 & 83.4 & 82.4 & 56.9 & 41.2 & 57.1 & 55.6 \\
\orangedot\hspace{2mm}Qwen2.5-14B & 64.6 & 48.7 & 41.3 & 48.7 & 35.3 & 62.9 & 33.1 & 81.2 & 75.4 & 56.2 & 47.5 & 58.6 & 52.6 \\
\orangesquare\hspace{2mm}Aya-Expanse-8B & 57.8 & 50.2 & 43.0 & 51.1 & \textbf{40.4} & 61.4 & 31.6 & 73.6 & 72.3 & 46.6 & 48.9 & 54.3 & 52.2 \\
\orangesquare\hspace{2mm}Qwen2.5-7B-Instruct & 61.8 & 47.1 & 41.1 & 46.7 & 32.2 & 61.3 & 30.3 & 73.4 & 71.2 & 47.6 & 52.0 & 54.3 & 50.9 \\
\orangedot\hspace{2mm}Qwen2.5-7B & 59.8 & 48.3 & 42.5 & 47.2 & 32.2 & 63.4 & 29.5 & 73.9 & 73.5 & 49.3 & 48.4 & 57.1 & 50.6 \\
\bluesquare\hspace{2mm}Llama-3.1-8B-Instruct & 60.8 & 48.5 & 40.9 & 44.6 & 34.3 & \textbf{65.7} & 32.3 & 70.8 & 75.5 & 38.1 & 41.6 & 64.3 & 50.1 \\
\bluesquare\hspace{2mm}Llama-3-8B-Instruct & 58.6 & 47.1 & 37.6 & 46.6 & 33.5 & 63.5 & 27.0 & 66.3 & 69.5 & 38.1 & 38.5 & 61.4 & 49.3 \\
\orangesquare\hspace{2mm}Aya-23-35B & 60.4 & 48.8 & 43.0 & 52.1 & 35.1 & 59.7 & 29.8 & 72.9 & 56.9 & 45.3 & 34.8 & 60.0 & 48.8 \\
\bluedot\hspace{2mm}Llama-3.1-8B & 62.6 & 47.6 & 38.8 & 46.0 & 35.1 & 63.2 & 31.3 & 61.4 & 54.1 & 30.6 & 32.1 & 58.6 & 45.8 \\
\orangesquare\hspace{2mm}Aya-23-8B & 59.6 & 49.3 & 42.1 & 48.9 & 37.3 & 62.7 & 27.0 & 60.7 & 45.0 & 33.0 & 34.4 & 58.6 & 45.7 \\
\orangesquare\hspace{2mm}Qwen2.5-3B-Instruct & 56.2 & 44.8 & 38.9 & 42.0 & 32.2 & 59.1 & 27.5 & 67.4 & 60.1 & 37.8 & 33.0 & 54.3 & 45.2 \\
\bluesquare\hspace{2mm}Mixtral-v0.1-7B-Instruct & 56.4 & 47.1 & 44.6 & 46.1 & 31.7 & 59.1 & 29.3 & 58.6 & 51.5 & 35.8 & 34.2 & 57.1 & 45.1 \\
\redsquare\hspace{2mm}Cere-Llama-3-8B & 60.2 & 48.7 & 41.8 & 46.9 & 35.9 & 63.1 & 28.0 & 51.4 & 48.1 & 25.6 & 33.9 & 58.6 & 44.8 \\
\bluedot\hspace{2mm}Llama-3-8B & 61.8 & 46.5 & 36.9 & 46.1 & 32.8 & 62.8 & 30.3 & 51.4 & 44.0 & 25.4 & 29.6 & 54.3 & 43.0 \\
\bluesquare\hspace{2mm}Ministral-2410-8B-Instruct & 57.4 & 45.3 & 37.5 & 44.1 & 31.9 & 60.6 & 31.3 & 60.9 & 40.5 & 26.4 & 24.9 & 58.6 & 42.8 \\
\orangesquare\hspace{2mm}Qwen2.5-5B-Instruct & 54.6 & 42.5 & 35.7 & 40.3 & 28.1 & 58.2 & 22.9 & 53.4 & 34.7 & 28.9 & 29.2 & 48.6 & 42.8 \\
\orangedot\hspace{2mm}Qwen2.5-3B & 55.2 & 43.9 & 38.0 & 40.6 & 29.7 & 59.5 & 26.5 & 61.9 & 43.5 & 22.6 & 24.4 & 55.7 & 41.5 \\
\redsquare\hspace{2mm}Kanarya-2B & 64.2 & 49.3 & 45.9 & 45.6 & 35.8 & 62.5 & 30.0 & 28.1 & 0.0 & 18.0 & 27.1 & 54.3 & 41.0 \\
\orangesquare\hspace{2mm}Aya-101-13B & 59.6 & 41.3 & 37.4 & 35.0 & 27.3 & 57.1 & 22.9 & 22.9 & 1.0 & 37.4 & 47.1 & 57.1 & 40.6 \\
\bluedot\hspace{2mm}Llama-3.2-3B & 57.0 & 45.4 & 40.0 & 43.2 & 31.5 & 59.5 & 29.5 & 47.3 & 19.9 & 29.0 & 29.0 & 57.1 & 40.6 \\
\orangedot\hspace{2mm}Qwen2.5-5B & 54.2 & 42.1 & 35.8 & 39.7 & 27.3 & 57.6 & 21.6 & 46.7 & 23.0 & 23.0 & 29.9 & 50.0 & 39.8 \\
\bluedot\hspace{2mm}Mistral-v0.1-7B & 56.6 & 45.2 & 42.8 & 39.5 & 29.2 & 60.2 & 24.2 & 37.4 & 30.8 & 20.3 & 25.3 & 57.1 & 39.2 \\
\bluesquare\hspace{2mm}Llama-3.2-3B-Instruct & 54.6 & 44.0 & 35.5 & 39.4 & 27.8 & 63.7 & 26.2 & 55.8 & 1.1 & 34.4 & 31.0 & 54.3 & 39.1 \\
\bluedot\hspace{2mm}Mistral-v0.3-7B & 58.4 & 43.4 & 41.0 & 38.6 & 26.6 & 58.4 & 24.2 & 41.1 & 27.6 & 26.9 & 23.5 & 57.1 & 39.0 \\
\bluesquare\hspace{2mm}Mistral-v0.3-7B-Instruct & 57.2 & 43.0 & 41.2 & 40.9 & 27.0 & 55.3 & 25.4 & 46.1 & 30.0 & 19.6 & 21.5 & 50.0 & 38.6 \\
\reddot\hspace{2mm}Trendyol-v1.0-7B-Base & 61.0 & 46.9 & 46.4 & 43.2 & 32.4 & 58.6 & 28.5 & 36.2 & 0.0 & 24.8 & 23.1 & 57.1 & 38.4 \\
\reddot\hspace{2mm}Commencis-7B & 58.0 & 41.3 & 34.8 & 38.6 & 27.6 & 56.5 & 24.7 & 32.3 & 22.7 & 24.7 & 24.2 & 58.6 & 38.1 \\
\orangedot\hspace{2mm}Qwen2.5-5B & 54.8 & 40.8 & 36.2 & 35.7 & 26.5 & 56.5 & 21.1 & 29.9 & 20.3 & 17.9 & 25.1 & 47.1 & 37.8 \\
\bluedot\hspace{2mm}Llama-3.2-1B & 55.6 & 42.1 & 36.2 & 37.3 & 29.2 & 57.7 & 28.5 & 29.6 & 21.7 & 18.9 & 22.4 & 52.9 & 37.7 \\
\orangesquare\hspace{2mm}Qwen2.5-5B-Instruct & 53.6 & 41.6 & 36.8 & 35.6 & 28.1 & 57.4 & 23.7 & 30.0 & 28.3 & 21.1 & 24.2 & 54.3 & 36.9 \\
\reddot\hspace{2mm}Turna-1B & 55.8 & 40.3 & 38.0 & 38.3 & 27.3 & 51.2 & 23.7 & 22.6 & 19.2 & 19.3 & 24.2 & 51.4 & 35.9 \\
\hline
\end{tabular}
}%
\caption{Multiple choice question answering results of the models, sorted by their average scores. Base and instruction-tuned variants are represented by circles and squares, respectively. Colors indicate the language focus: blue for English-focused, yellow for multilingual-focused, and red for Turkish-focused models.}
\label{appendix_mcqa}
\end{table*}
}

%% file: tables/mt.tex
\renewcommand{\arraystretch}{1.05}{
\begin{table*}[ht]
\centering
\begin{tabular}{lrr}
\toprule
\textbf{} & \textbf{WMT16EN-TR} & \textbf{Avg.} \\
\hline
\orangesquare\hspace{2mm}Aya-Expanse-32B & \textbf{20.1} & \textbf{20.1} \\
\orangesquare\hspace{2mm}Aya-23-35B & 18.5 & 18.5 \\
\orangesquare\hspace{2mm}Aya-Expanse-8B & 17.0 & 17.0 \\
\orangesquare\hspace{2mm}Aya-23-8B & 16.9 & 16.9 \\
\bluesquare\hspace{2mm}Llama-3.1-8B-Instruct & 15.6 & 15.6 \\
\bluedot\hspace{2mm}Llama-3.1-8B & 15.6 & 15.6 \\
\bluesquare\hspace{2mm}Llama-3.3-70B-Instruct & 13.6 & 13.6 \\
\bluesquare\hspace{2mm}Llama-3-8B-Instruct & 13.6 & 13.6 \\
\orangesquare\hspace{2mm}Qwen2.5-14B-Instruct & 12.4 & 12.4 \\
\bluedot\hspace{2mm}Llama-3-8B & 11.3 & 11.3 \\
\bluesquare\hspace{2mm}Ministral-2410-8B-Instruct & 11.2 & 11.2 \\
\bluesquare\hspace{2mm}Mixtral-v0.1-7B-Instruct & 9.5 & 9.5 \\
\orangedot\hspace{2mm}Qwen2.5-14B & 8.1 & 8.1 \\
\bluedot\hspace{2mm}Llama-3.2-3B & 7.9 & 7.9 \\
\orangesquare\hspace{2mm}Qwen2.5-3B-Instruct & 6.6 & 6.6 \\
\bluesquare\hspace{2mm}Mistral-v0.3-7B-Instruct & 5.9 & 5.9 \\
\orangedot\hspace{2mm}Qwen2.5-7B & 5.5 & 5.5 \\
\orangesquare\hspace{2mm}Qwen2.5-7B-Instruct & 5.4 & 5.4 \\
\bluedot\hspace{2mm}Mistral-v0.1-7B & 5.0 & 5.0 \\
\bluesquare\hspace{2mm}Llama-3.2-3B-Instruct & 4.9 & 4.9 \\
\reddot\hspace{2mm}Commencis-7B & 4.7 & 4.7 \\
\orangesquare\hspace{2mm}Qwen2.5-5B-Instruct & 4.7 & 4.7 \\
\orangedot\hspace{2mm}Qwen2.5-3B & 4.1 & 4.1 \\
\orangedot\hspace{2mm}Qwen2.5-5B & 4.1 & 4.1 \\
\bluedot\hspace{2mm}Mistral-v0.3-7B & 3.8 & 3.8 \\
\orangesquare\hspace{2mm}Qwen2.5-5B-Instruct & 3.1 & 3.1 \\
\redsquare\hspace{2mm}Kanarya-2B & 3.0 & 3.0 \\
\orangedot\hspace{2mm}Qwen2.5-5B & 1.4 & 1.4 \\
\bluedot\hspace{2mm}Llama-3.2-1B & 1.2 & 1.2 \\
\reddot\hspace{2mm}Trendyol-v1.0-7B-Base & 1.0 & 1.0 \\
\redsquare\hspace{2mm}Cere-Llama-3-8B & 0.1 & 0.1 \\
\reddot\hspace{2mm}Turna-1B & 0.1 & 0.1 \\
\orangesquare\hspace{2mm}Aya-101-13B & 0.0 & 0.0 \\
\hline
\end{tabular}
\caption{Machine translation results of the models, sorted by their average scores. Base and instruction-tuned variants are represented by circles and squares, respectively. Colors indicate the language focus: blue for English-focused, yellow for multilingual-focused, and red for Turkish-focused models.}
\label{appendix_mt}
\end{table*}
}

%% file: tables/nli.tex
\renewcommand{\arraystretch}{1.05}{
\begin{table*}[ht]
\centering
\begin{tabular}{lrrrr}
\toprule
\textbf{} & \textbf{MNLI} & \textbf{SNLI} & \textbf{XNLI} & \textbf{Avg.} \\
\hline
\orangesquare\hspace{2mm}Aya-Expanse-32B & \textbf{42.5} & \textbf{47.1} & \textbf{40.5} & \textbf{43.3} \\
\orangesquare\hspace{2mm}Aya-23-35B & 37.3 & 37.3 & 38.1 & 37.6 \\
\bluesquare\hspace{2mm}Llama-3.1-8B-Instruct & 36.7 & 35.9 & 35.4 & 36.0 \\
\orangesquare\hspace{2mm}Qwen2.5-14B-Instruct & 35.3 & 36.3 & 34.1 & 35.2 \\
\orangesquare\hspace{2mm}Qwen2.5-7B-Instruct & 35.6 & 35.5 & 33.8 & 35.0 \\
\orangesquare\hspace{2mm}Aya-Expanse-8B & 36.1 & 31.6 & 36.3 & 34.7 \\
\reddot\hspace{2mm}Trendyol-v1.0-7B-Base & 35.2 & 33.7 & 34.0 & 34.3 \\
\bluedot\hspace{2mm}Llama-3.2-1B & 35.0 & 33.7 & 33.5 & 34.1 \\
\reddot\hspace{2mm}Turna-1B & 34.9 & 33.8 & 33.4 & 34.1 \\
\bluesquare\hspace{2mm}Llama-3.2-3B-Instruct & 34.8 & 33.8 & 33.4 & 34.0 \\
\bluedot\hspace{2mm}Mistral-v0.3-7B & 34.9 & 33.4 & 33.6 & 34.0 \\
\orangesquare\hspace{2mm}Qwen2.5-3B-Instruct & 34.8 & 33.7 & 33.4 & 34.0 \\
\orangedot\hspace{2mm}Qwen2.5-14B & 34.8 & 33.7 & 33.4 & 34.0 \\
\bluesquare\hspace{2mm}Ministral-2410-8B-Instruct & 34.8 & 33.7 & 33.4 & 34.0 \\
\redsquare\hspace{2mm}Cere-Llama-3-8B & 34.9 & 33.7 & 33.3 & 34.0 \\
\orangedot\hspace{2mm}Qwen2.5-3B & 34.8 & 33.7 & 33.4 & 34.0 \\
\orangedot\hspace{2mm}Qwen2.5-7B & 34.8 & 33.6 & 33.4 & 34.0 \\
\orangesquare\hspace{2mm}Qwen2.5-5B-Instruct & 34.8 & 33.7 & 33.3 & 33.9 \\
\bluedot\hspace{2mm}Mistral-v0.1-7B & 34.8 & 33.6 & 33.4 & 33.9 \\
\orangesquare\hspace{2mm}Qwen2.5-5B-Instruct & 34.8 & 33.7 & 33.3 & 33.9 \\
\orangedot\hspace{2mm}Qwen2.5-5B & 34.8 & 33.7 & 33.3 & 33.9 \\
\bluedot\hspace{2mm}Llama-3-8B & 34.8 & 33.6 & 33.4 & 33.9 \\
\orangedot\hspace{2mm}Qwen2.5-5B & 34.7 & 33.7 & 33.4 & 33.9 \\
\bluesquare\hspace{2mm}Llama-3-8B-Instruct & 35.1 & 33.3 & 33.1 & 33.8 \\
\bluesquare\hspace{2mm}Mixtral-v0.1-7B-Instruct & 34.8 & 32.2 & 34.1 & 33.7 \\
\orangesquare\hspace{2mm}Aya-23-8B & 32.7 & 33.6 & 33.7 & 33.3 \\
\redsquare\hspace{2mm}Kanarya-2B & 33.4 & 31.7 & 34.2 & 33.1 \\
\bluedot\hspace{2mm}Llama-3.2-3B & 32.0 & 32.5 & 33.0 & 32.5 \\
\reddot\hspace{2mm}Commencis-7B & 33.3 & 31.5 & 32.7 & 32.5 \\
\bluesquare\hspace{2mm}Llama-3.3-70B-Instruct & 32.1 & 31.7 & 33.3 & 32.4 \\
\bluedot\hspace{2mm}Llama-3.1-8B & 31.9 & 34.3 & 30.3 & 32.2 \\
\bluesquare\hspace{2mm}Mistral-v0.3-7B-Instruct & 31.3 & 31.8 & 32.0 & 31.7 \\
\orangesquare\hspace{2mm}Aya-101-13B & 27.9 & 25.6 & 30.0 & 27.8 \\
\hline
\end{tabular}
\caption{Natural language inference results of the models, sorted by their average scores. Base and instruction-tuned variants are represented by circles and squares, respectively. Colors indicate the language focus: blue for English-focused, yellow for multilingual-focused, and red for Turkish-focused models.}
\label{appendix_nli}
\end{table*}
}

%% file: tables/qa.tex
\renewcommand{\arraystretch}{1.05}{
\begin{table*}[ht]
\centering
\begin{tabular}{lrrrr}
\toprule
\textbf{} & \textbf{XQUAD} & \textbf{TQUAD} & \textbf{MKQA} & \textbf{Avg.} \\
\hline
\orangedot\hspace{2mm}Qwen2.5-14B & \textbf{40.3} & 34.8 & 5.0 & \textbf{26.7} \\
\orangesquare\hspace{2mm}Aya-Expanse-32B & 31.9 & 30.2 & 16.4 & 26.2 \\
\redsquare\hspace{2mm}Cere-Llama-3-8B & 21.2 & \textbf{49.2} & 2.2 & 24.2 \\
\orangesquare\hspace{2mm}Aya-23-35B & 30.9 & 20.6 & \textbf{19.4} & 23.7 \\
\orangedot\hspace{2mm}Qwen2.5-5B & 31.8 & 34.3 & 0.3 & 22.1 \\
\bluedot\hspace{2mm}Llama-3-8B & 20.8 & 28.5 & 13.5 & 20.9 \\
\orangedot\hspace{2mm}Qwen2.5-7B & 31.9 & 28.0 & 1.4 & 20.5 \\
\orangedot\hspace{2mm}Qwen2.5-3B & 32.3 & 26.8 & 0.1 & 19.7 \\
\bluedot\hspace{2mm}Llama-3.1-8B & 20.9 & 27.6 & 9.2 & 19.3 \\
\orangesquare\hspace{2mm}Aya-23-8B & 24.7 & 20.6 & 10.0 & 18.5 \\
\bluesquare\hspace{2mm}Llama-3.1-8B-Instruct & 21.4 & 23.3 & 9.1 & 18.0 \\
\bluedot\hspace{2mm}Mistral-v0.3-7B & 17.1 & 21.9 & 11.5 & 16.8 \\
\bluedot\hspace{2mm}Mistral-v0.1-7B & 16.7 & 21.0 & 12.0 & 16.6 \\
\bluesquare\hspace{2mm}Llama-3.3-70B-Instruct & 14.5 & 17.4 & 16.3 & 16.1 \\
\bluesquare\hspace{2mm}Llama-3.2-3B-Instruct & 23.0 & 18.7 & 4.7 & 15.5 \\
\orangesquare\hspace{2mm}Aya-Expanse-8B & 25.0 & 13.5 & 7.0 & 15.1 \\
\bluedot\hspace{2mm}Llama-3.2-3B & 15.5 & 21.2 & 6.0 & 14.2 \\
\bluesquare\hspace{2mm}Ministral-2410-8B-Instruct & 22.9 & 17.6 & 2.2 & 14.2 \\
\bluesquare\hspace{2mm}Mixtral-v0.1-7B-Instruct & 10.7 & 9.8 & 12.5 & 11.0 \\
\orangesquare\hspace{2mm}Qwen2.5-5B-Instruct & 16.5 & 14.7 & 0.2 & 10.4 \\
\orangesquare\hspace{2mm}Qwen2.5-5B-Instruct & 13.4 & 16.3 & 0.7 & 10.1 \\
\bluesquare\hspace{2mm}Llama-3-8B-Instruct & 9.7 & 12.9 & 4.2 & 9.0 \\
\bluesquare\hspace{2mm}Mistral-v0.3-7B-Instruct & 11.4 & 9.5 & 5.0 & 8.7 \\
\orangesquare\hspace{2mm}Aya-101-13B & 7.6 & 5.4 & 2.5 & 5.2 \\
\reddot\hspace{2mm}Commencis-7B & 6.6 & 5.4 & 3.2 & 5.1 \\
\bluedot\hspace{2mm}Llama-3.2-1B & 4.9 & 6.3 & 0.8 & 4.0 \\
\orangedot\hspace{2mm}Qwen2.5-5B & 4.0 & 7.2 & 0.1 & 3.8 \\
\orangesquare\hspace{2mm}Qwen2.5-3B-Instruct & 6.1 & 3.3 & 0.1 & 3.1 \\
\redsquare\hspace{2mm}Kanarya-2B & 0.8 & 1.7 & 0.6 & 1.0 \\
\orangesquare\hspace{2mm}Qwen2.5-7B-Instruct & 0.9 & 0.6 & 0.0 & 0.5 \\
\orangesquare\hspace{2mm}Qwen2.5-14B-Instruct & 0.9 & 0.3 & 0.0 & 0.4 \\
\reddot\hspace{2mm}Trendyol-v1.0-7B-Base & 0.0 & 0.8 & 0.1 & 0.3 \\
\reddot\hspace{2mm}Turna-1B & 0.0 & 0.0 & 0.1 & 0.0 \\
\hline
\end{tabular}
\caption{Open ended question answering results results of the models, sorted by their average scores. Base and instruction-tuned variants are represented by circles and squares, respectively. Colors indicate the language focus: blue for English-focused, yellow for multilingual-focused, and red for Turkish-focused models.}
\label{appendix_gec}
\end{table*}
}

%% file: tables/sum.tex
\renewcommand{\arraystretch}{1.05}{
\begin{table*}[ht]
\centering
\begin{tabular}{lrrrrr}
\toprule
\textbf{} & \textbf{XLSum} & \textbf{WikiLingua} & \textbf{WikiHowSumm} & \textbf{MLSum} & \textbf{Avg.} \\
\hline
\orangesquare\hspace{2mm}Aya-Expanse-32B & \textbf{21.4} & \textbf{21.5} & \textbf{16.6} & \textbf{30.1} & \textbf{22.4} \\
\orangesquare\hspace{2mm}Aya-23-35B & 13.3 & 18.4 & 12.7 & 25.9 & 17.6 \\
\bluesquare\hspace{2mm}Llama-3.3-70B-Instruct & 16.3 & 12.0 & 8.5 & 28.1 & 16.2 \\
\bluesquare\hspace{2mm}Llama-3-8B-Instruct & 13.5 & 8.1 & 7.9 & 25.8 & 13.8 \\
\bluesquare\hspace{2mm}Llama-3.1-8B-Instruct & 12.4 & 7.9 & 7.9 & 25.8 & 13.5 \\
\bluedot\hspace{2mm}Llama-3.1-8B & 12.4 & 7.9 & 7.9 & 25.8 & 13.5 \\
\orangedot\hspace{2mm}Qwen2.5-14B & 13.1 & 6.8 & 6.6 & 25.7 & 13.0 \\
\bluesquare\hspace{2mm}Ministral-2410-8B-Instruct & 13.8 & 6.9 & 6.7 & 24.0 & 12.8 \\
\orangesquare\hspace{2mm}Qwen2.5-14B-Instruct & 15.7 & 7.1 & 4.7 & 23.7 & 12.8 \\
\orangedot\hspace{2mm}Qwen2.5-7B & 12.3 & 6.9 & 6.6 & 25.3 & 12.8 \\
\orangesquare\hspace{2mm}Aya-23-8B & 14.1 & 11.4 & 1.7 & 24.0 & 12.8 \\
\bluedot\hspace{2mm}Llama-3-8B & 11.1 & 6.6 & 6.6 & 25.0 & 12.3 \\
\orangedot\hspace{2mm}Qwen2.5-3B & 11.4 & 6.5 & 6.5 & 24.9 & 12.3 \\
\orangesquare\hspace{2mm}Aya-Expanse-8B & 13.3 & 13.4 & 0.5 & 21.4 & 12.2 \\
\bluedot\hspace{2mm}Llama-3.2-3B & 12.0 & 6.4 & 6.5 & 23.3 & 12.1 \\
\orangesquare\hspace{2mm}Qwen2.5-3B-Instruct & 12.0 & 6.6 & 6.8 & 22.7 & 12.0 \\
\orangesquare\hspace{2mm}Qwen2.5-5B-Instruct & 11.0 & 6.1 & 6.3 & 24.6 & 12.0 \\
\orangedot\hspace{2mm}Qwen2.5-5B & 11.9 & 5.9 & 6.3 & 24.0 & 12.0 \\
\orangedot\hspace{2mm}Qwen2.5-5B & 9.2 & 5.6 & 6.1 & 24.3 & 11.3 \\
\bluesquare\hspace{2mm}Mistral-v0.3-7B-Instruct & 10.9 & 6.2 & 3.7 & 24.1 & 11.2 \\
\orangesquare\hspace{2mm}Qwen2.5-7B-Instruct & 11.6 & 6.3 & 5.8 & 20.5 & 11.0 \\
\redsquare\hspace{2mm}Kanarya-2B & 9.3 & 4.5 & 5.3 & 24.7 & 10.9 \\
\bluesquare\hspace{2mm}Mixtral-v0.1-7B-Instruct & 10.6 & 6.0 & 4.9 & 22.2 & 10.9 \\
\orangesquare\hspace{2mm}Qwen2.5-5B-Instruct & 9.0 & 5.0 & 5.8 & 22.9 & 10.7 \\
\reddot\hspace{2mm}Commencis-7B & 9.5 & 6.3 & 7.2 & 16.1 & 9.8 \\
\bluesquare\hspace{2mm}Llama-3.2-3B-Instruct & 6.7 & 4.1 & 6.6 & 14.8 & 8.1 \\
\bluedot\hspace{2mm}Mistral-v0.3-7B & 6.3 & 6.0 & 1.9 & 17.3 & 7.9 \\
\bluedot\hspace{2mm}Llama-3.2-1B & 7.2 & 3.5 & 5.9 & 13.4 & 7.5 \\
\reddot\hspace{2mm}Turna-1B & 6.1 & 5.6 & 5.7 & 11.0 & 7.1 \\
\reddot\hspace{2mm}Trendyol-v1.0-7B-Base & 5.0 & 2.9 & 4.6 & 8.8 & 5.3 \\
\bluedot\hspace{2mm}Mistral-v0.1-7B & 1.2 & 6.0 & 1.1 & 6.3 & 3.6 \\
\redsquare\hspace{2mm}Cere-Llama-3-8B & 0.6 & 2.1 & 5.2 & 6.0 & 3.5 \\
\orangesquare\hspace{2mm}Aya-101-13B & 0.1 & 0.0 & 0.0 & 0.3 & 0.1 \\
\hline
\end{tabular}
\caption{Summarization results of the models, sorted by their average scores. Shapes represent the model architectures: an inverted triangle for Lllama, a star for Aya, a circle for Qwen, a triangle for T5, and a pentagon for GPT-J. Colors indicate the language focus of the models: blue for English-focused, yellow for multilingual-focused, and red for Turkish-focused models.}
\label{appendix_sum}
\end{table*}
}

%% file: tables/tc.tex
\renewcommand{\arraystretch}{1.05}{
\begin{table*}[ht]
\centering
\begin{tabular}{lrrrrr}
\toprule
\textbf{} & \textbf{STSb} & \textbf{OffensEval} & \textbf{NewsCat} & \textbf{IronyTR} & \textbf{Avg.} \\
\hline
\bluesquare\hspace{2mm}Llama-3.3-70B-Instruct & 12.9 & \textbf{83.1} & 78.0 & 58.2 & \textbf{58.1} \\
\orangesquare\hspace{2mm}Aya-Expanse-32B & 21.5 & 67.1 & \textbf{82.8} & 50.0 & 55.3 \\
\orangedot\hspace{2mm}Qwen2.5-7B & 17.0 & 77.4 & 54.8 & 57.3 & 51.6 \\
\orangesquare\hspace{2mm}Qwen2.5-7B-Instruct & 18.3 & 80.3 & 40.0 & 55.0 & 48.4 \\
\redsquare\hspace{2mm}Kanarya-2B & 12.9 & 61.6 & 66.8 & 50.0 & 47.8 \\
\bluesquare\hspace{2mm}Mixtral-v0.1-7B-Instruct & 13.0 & 62.9 & 54.0 & 52.5 & 45.6 \\
\orangesquare\hspace{2mm}Aya-23-8B & 23.0 & 34.2 & 72.4 & 51.7 & 45.3 \\
\bluedot\hspace{2mm}Llama-3.1-8B & 17.0 & 34.6 & 69.2 & 58.5 & 44.8 \\
\redsquare\hspace{2mm}Cere-Llama-3-8B & 22.1 & 34.0 & 68.4 & 50.2 & 43.7 \\
\bluedot\hspace{2mm}Llama-3.2-1B & 17.1 & 46.7 & 58.0 & 52.8 & 43.7 \\
\orangesquare\hspace{2mm}Aya-Expanse-8B & 21.0 & 26.8 & 76.0 & 50.5 & 43.6 \\
\orangesquare\hspace{2mm}Qwen2.5-14B-Instruct & 24.9 & 54.7 & 32.4 & \textbf{61.3} & 43.3 \\
\bluesquare\hspace{2mm}Mistral-v0.3-7B-Instruct & 12.9 & 45.2 & 61.2 & 50.7 & 42.5 \\
\orangesquare\hspace{2mm}Aya-101-13B & 17.0 & 79.9 & 20.0 & 52.2 & 42.3 \\
\reddot\hspace{2mm}Trendyol-v1.0-7B-Base & 15.5 & 20.3 & 81.2 & 50.0 & 41.8 \\
\bluesquare\hspace{2mm}Llama-3-8B-Instruct & 14.2 & 30.8 & 62.8 & 54.5 & 40.6 \\
\bluedot\hspace{2mm}Llama-3-8B & 16.4 & 21.9 & 72.4 & 51.5 & 40.6 \\
\orangedot\hspace{2mm}Qwen2.5-3B & 12.9 & 48.4 & 44.8 & 54.7 & 40.2 \\
\bluesquare\hspace{2mm}Llama-3.1-8B-Instruct & 19.6 & 23.6 & 66.0 & 51.3 & 40.1 \\
\reddot\hspace{2mm}Commencis-7B & 14.9 & 24.3 & 62.4 & 56.0 & 39.4 \\
\bluedot\hspace{2mm}Llama-3.2-3B & 13.2 & 25.4 & 66.4 & 50.0 & 38.7 \\
\orangesquare\hspace{2mm}Aya-23-35B & \textbf{25.4} & 21.0 & 55.6 & 50.2 & 38.0 \\
\bluesquare\hspace{2mm}Ministral-2410-8B-Instruct & 21.4 & 20.3 & 60.4 & 50.0 & 38.0 \\
\orangedot\hspace{2mm}Qwen2.5-14B & 20.4 & 22.0 & 52.4 & 56.2 & 37.7 \\
\bluedot\hspace{2mm}Mistral-v0.3-7B & 14.2 & 20.7 & 66.0 & 49.8 & 37.7 \\
\bluedot\hspace{2mm}Mistral-v0.1-7B & 13.6 & 20.5 & 65.2 & 50.2 & 37.3 \\
\bluesquare\hspace{2mm}Llama-3.2-3B-Instruct & 12.9 & 20.6 & 64.0 & 50.2 & 36.9 \\
\reddot\hspace{2mm}Turna-1B & 14.2 & 51.0 & 32.8 & 48.3 & 36.6 \\
\orangesquare\hspace{2mm}Qwen2.5-3B-Instruct & 16.8 & 37.6 & 37.2 & 51.3 & 35.7 \\
\orangedot\hspace{2mm}Qwen2.5-5B & 12.9 & 27.4 & 48.4 & 52.3 & 35.3 \\
\orangesquare\hspace{2mm}Qwen2.5-5B-Instruct & 12.9 & 20.7 & 48.8 & 49.7 & 33.0 \\
\orangedot\hspace{2mm}Qwen2.5-5B & 12.9 & 33.7 & 26.8 & 49.3 & 30.7 \\
\orangesquare\hspace{2mm}Qwen2.5-5B-Instruct & 13.1 & 21.4 & 29.2 & 47.3 & 27.8 \\
\hline
\end{tabular}
\caption{Text classification results of the models, sorted by their average scores. Shapes represent the model architectures: an inverted triangle for Lllama, a star for Aya, a circle for Qwen, a triangle for T5, and a pentagon for GPT-J. Colors indicate the language focus of the models: blue for English-focused, yellow for multilingual-focused, and red for Turkish-focused models.}
\label{appendix_tc}
\end{table*}
}

%% file: sections/sec_appendix_samples.tex
\input{sections/lira}
\begin{question}[Belebele]
Tüm notalara doğru şekilde basmaya devam ederken elinizin mümkün olduğu kadar rahat olduğundan emin olun - aynı zamanda parmaklarınızla fazladan hareketler yapmamaya çalışın. Bu şekilde kendinizi olabildiğince az yormuş olacaksınız. Unutmayın ki piyanoda olduğu gibi daha fazla ses için tuşlara çok güçlü vurmanıza gerek yoktur. Akordeon üzerinde, ekstra hacim elde etmek için körüğü daha fazla basınç veya hızda kullanırsınız.

Metne göre, hangisi akordeonu başarılı bir şekilde çalmak için uygun bir tavsiye değildir?
\begin{choices}
  \correctchoice {Daha fazla ses çıkarmak için tuşlara daha güçlü basın}
  \choice Yorulmamak için gereksiz hareketleri en aza indirin
  \choice Eliniz rahat pozisyondayken notalara doğru şekilde basın
  \choice Ekstra ses elde etmek için körüğü daha hızlı kullanın
\end{choices}
\DashedLineWithText{English Translation}\\
While continuing to press all the notes correctly, make sure your hand is as relaxed as possible – at the same time, try not to make extra movements with your fingers. This way, you will tire yourself as little as possible. Remember that, just like on the piano, you don’t need to hit the keys very hard to produce more sound. On the accordion, to achieve extra volume, you use the bellows with more pressure or speed.

According to the text, which of the following is *not* an appropriate piece of advice for playing the accordion successfully?
\begin{choices}
  \correctchoice {Press the keys harder to produce more sound}
  \choice Minimize unnecessary movements to avoid fatigue
  \choice Press the notes correctly while keeping your hand relaxed
  \choice Use the bellows faster to achieve extra volume
\end{choices}
\end{question}

\begin{question}[BilmeceBench]
Bilmece: Kuyruklu kumbara yemek taşır ambara.

Bilmecenin anlamı aşağıdakilerden hangisidir?
\begin{choices}
  \choice BALTA
  \choice ARMUT
  \correctchoice {KAŞIK}
  \choice AYAKKABI
\end{choices}
\DashedLineWithText{English Translation}\\
Riddle: A piggy bank with a tail carries food to the storage.

What is the meaning of the riddle?
\begin{choices}
  \choice AXE
  \choice PEAR
  \correctchoice {SPOON}
  \choice SHOE
\end{choices}
\end{question}

\begin{question}[Circumflex]
Kelime: Hakim

Kelimenin anlamı aşağıdakilerden hangisidir?

Cevap:
\begin{choices}
  \correctchoice {Sıfat: Egemenliğini yürüten, buyruğunu yürüten, sözünü geçiren}
  \choice Sıfat: Bilge
\end{choices}
\DashedLineWithText{English Translation}
Word: Hakim
What is the meaning of the word?
Answer:
\begin{choices}
  \correctchoice {Adjective: One who exercises authority, enforces command, and has influence}
  \choice Adjective: Wise
\end{choices}
\end{question}

\begin{question}[Exams]
Glikokortikoidler olarak adlandırılan hormonlar nerede sentezlenirler:
\begin{choices}
  \choice tiroid bezinde.
  \choice hipofizde.
  \choice pankreasta.
  \correctchoice {böbrek üstü bezinin kabuğunda.}
\end{choices}
\DashedLineWithText{English Translation}
Where are the hormones called glucocorticoids synthesized:
\begin{choices}
  \choice in the thyroid gland.
  \choice in the pituitary gland.
  \choice in the pancreas.
  \correctchoice {in the cortex of the adrenal gland.}
\end{choices}
\end{question}

\begin{question}[IronyTR]
Cümle: *ODTÜden mezun olmadan yapılacak 100 şey* Madde 101: Pikapla kampüsten kaçmak

Soru: Bu cümlede ironi var mı?
\begin{choices}
  \choice Hayır
  \correctchoice {Evet}
\end{choices}
\DashedLineWithText{English Translation}
Sentence: *100 things to do before graduating from METU* Item 101: Escape the campus with a pickup truck
Question: Is there irony in this sentence?
\begin{choices}
  \choice No
  \correctchoice {Yes}
\end{choices}
\end{question}

\begin{question}[Natural Language Inference]
Aşağıda iki cümle verilmektedir:
\\
Cümle 1: "Evet, sanırım en sevdiğim restoran her zaman en yakın restorandır. En yakın olanı biliyorsun. En düşük kriterlere uyduğu sürece."

Cümle 2: "En sevdiğim restoranlar her zaman evimden en az yüz mil uzakta."
\\
\\
Bu iki cümle arasındaki ilişki nedir:
\begin{choices}
  \choice TUTARLI
  \choice ALAKASIZ
  \correctchoice {ÇELİŞKİLİ}
\end{choices}
\DashedLineWithText{English Translation}
\\
Below are two sentences:
\\
\\
Sentence 1: "Yes, I guess my favorite restaurant is always the closest one. You know the closest. As long as it meets the lowest standards."
\\
Sentence 2: "My favorite restaurants are always at least a hundred miles away from my home."
\\
\\
What is the relationship between these two sentences?
\begin{choices}
  \choice ENTAILMENT
  \choice NEUTRAL
  \correctchoice {CONTRADICTION}
\end{choices}
\end{question}

\begin{question}[NewsCat]
Cümle: Hırsız, Hietanen'in başını yaktı DENİZLİSPORLU Hietanen'i hırsız yaktı. Porto maçı için kampta olduğu saatte evine giren hırsız, yatak odasına geçip, içki içti. Eşi Riiena eve döndüğünde yatağı dağınık görünce, "Bana kampta olduğunu söylüyorsun, eve kadın getiriyorsun" diyerek ayrılmak istedi. Rıza Çalımbay'dan izin alan futbolcu, eşini ikna etti. \\
...\\
Soru: Bu cümlenin konusu nedir?\\
Cevap:
\begin{choices}
  \correctchoice {spor}
  \choice magazin
  \choice siyaset
  \choice sağlık
  \choice ekonomi
\end{choices}
\DashedLineWithText{English Translation}
\\
Sentence: The thief got Hietanen in trouble. The thief caused trouble for Denizlispor player Hietanen. While he was at camp for the Porto match, a thief entered his home, went into the bedroom, and drank alcohol. When his wife Riiena returned home and saw the bed messy, she said, "You tell me you're at camp, but you bring a woman home," and wanted to leave him. The footballer got permission from Rıza Çalımbay and convinced his wife.
\\
...
\\
Question: What is the topic of this sentence?
\\
Answer:
\begin{choices}
  \correctchoice {sports}
  \choice celebrity news
  \choice politics
  \choice health
  \choice economy
\end{choices}
\end{question}

\begin{question}[OffenseEval]
Cümle: Hala Hogwarts mektubum gelmediğinden oluyor tüm bunlar.
\\
Soru: Bu cümle nefret söylemi içermekte midir?
\begin{choices}
  \correctchoice {Hayır}
  \choice Evet
\end{choices}
\DashedLineWithText{English Translation}
\\
Sentence: All of this is happening because I still haven’t received my Hogwarts letter.
\\
Question: Does this sentence contain hate speech?
\begin{choices}
  \correctchoice {No}
  \choice Yes
\end{choices}
\end{question}

\begin{question}[STSb]
Aşağıda iki cümle verilmektedir:
\\
\\
Cümle 1: "Bir kız saçlarını şekillendirmekte."\\
Cümle 2: "Bir kız saçını fırçalıyor."\\\\
Bu iki cümle arasında ne kadar benzerlik vardır:
\begin{choices}
  \choice Benzerlik Yok
  \choice Düşük Benzerlik
  \choice Orta Benzerlik
  \correctchoice {Yüksek Benzerlik}
  \choice Çok Yüksek Benzerlik
  \choice Mantıksal Olarak Aynı
\end{choices}
\DashedLineWithText{English Translation}\\
Below are two sentences:
\\
\\
Sentence 1: "A girl is styling her hair."\\
Sentence 2: "A girl is brushing her hair."\\\\
How similar are these two sentences?
\begin{choices}
  \choice No Similarity
  \choice Low Similarity
  \choice Moderate Similarity
  \correctchoice {High Similarity}
  \choice Very High Similarity
  \choice Logically Equivalent
\end{choices}
\end{question}
\begin{question}[TQUAD]
Kaynak: Kemaleddin ibn Yunus ya da Musa ibn Yunus (doğum yılı ve yeri: 1156 Musul - ölüm yılı ve yeri: 1241 Musul).Astronom, matematikçi ve İslam bilgini.Tam adı Musa bin Yunus bin Muhammed bin Men’a'dır, Künyesi ise Ebu’l-Feth’tir, lakabı Kemaleddin olup ayrıca İbn-i Yunus ve Mewsilî diye de bilinir.İlk eğitimini babası Şeyh Yunus Rızauddin'in yanında fıkıh ve hadis ilimleri öğrendi, ardından Bağdat'taki Nizamiye Medreseleri'nde okumaya devam etti. Burada Şerafeddin el-Tusî'den matematik derslerini aldı, ardından Batlamyus'un Almagest adlı eserini de öğrenir. Ardından Musul'a döndü, Emir Zeyneddin Camii'nde dersler verdi. İlim öğretmeye elverişli olarak inşa edilen bu cami Kemaliyye Medresesi olarak anıldı. Kısa zamanda şöhreti etrafa yayılan Musa Kemaleddin ibn Yunus pek çok çevreden gelen talebelere ilim öğretti.\\\\
Soru: Kemaleddin ibn Yunus lakabı dışında hangi isimlerle bilinir?\\\\
Cevap:\\
\textcolor{ForestGreen}{İbn-i Yunus ve Mewsilî}
\\\DashedLineWithText{English Translation}\\
Source: Kemaleddin ibn Yunus or Musa ibn Yunus (born 1156 Mosul – died 1241 Mosul).
Astronomer, mathematician, and Islamic scholar.
His full name is Musa bin Yunus bin Muhammad bin Men’a; his kunyah is Abu’l-Feth, his laqab is Kemaleddin, and he is also known as İbn-i Yunus and Mewsilî.
He received his initial education in fiqh and hadith from his father, Sheikh Yunus Rızauddin,
then continued at the Nizāmiyya Madrasas in Baghdad, studying mathematics under Sharaf al-Din al-Tusi and learning Ptolemy’s Almagest.
He returned to Mosul and taught at the Zayn al-Dīn Mosque, also called the Kemaliyye Madrasa, gaining fame and attracting many students.

Question: Aside from the laqab Kemaleddin ibn Yunus, by what other names is he known?  
Answer:\\
\textcolor{ForestGreen}{İbn-i Yunus ve Mewsilî}
\end{question}

\begin{question}[Turkce Atasozleri]
Atasözü: aba altında er yatar\\
Yukarıdaki atasözünün tanımı aşağıdakilerden hangisidir?
\begin{choices}
  \correctchoice {Giyim kuşam kişiliğe ölçü olamaz.}
  \choice Tanrı'dan korkmayan kimse, insana her türlü kötülüğü yapabilir.
  \choice İnsan kendinde herhangi bir kusur varken başkalarını aynı kusurla suçlamamalıdır.
  \choice Ortaya çıkan bir yanlışlık çok geç de olsa düzeltilebilir.
\end{choices}
\DashedLineWithText{English Translation}\\
Proverb: Beneath a coarse cloak may lie a noble man  
What is the meaning of the proverb above?
\begin{choices}
  \correctchoice {Clothing and appearance are not reliable measures of character.}
  \choice One who does not fear God is capable of doing all kinds of harm to others.
  \choice A person should not accuse others of a fault they themselves possess.
  \choice A mistake that has come to light can be corrected, even if belatedly.
\end{choices}
\end{question}

\begin{question}[TurkishPLU / Goal Inference]
Örnek Adım: İşletme adının hemen altındaki ekranın sağ tarafında bulunan "Yer İşareti" düğmesine dokunun. Hedef:
\begin{choices}
  \choice Yelp'e İşletme Fotoğrafı Eklemek
  \choice Audacity'de İz İşaretleri Eklemek
  \correctchoice {Yelp'te Bir İşletmeye Yer İşareti Eklemek}
  \choice Yelp'te Yinelenen İşletme Girişlerini Bildirmek
\end{choices}
\DashedLineWithText{English Translation}\\
Example Step: Tap the "Bookmark" button located on the right side of the screen just below the business name.  
Goal:
\begin{choices}
  \choice Add a Business Photo on Yelp
  \choice Add Track Markers in Audacity
  \correctchoice {Bookmark a Business on Yelp}
  \choice Report Duplicate Business Listings on Yelp
\end{choices}
\end{question}

\begin{question}[TurkishPLU / Next Event Prediction]
Hedef: Dâhilî Numara Nasıl Aranır? Adım: Arama cevaplandığı anda dâhilî numarayı gireceksen bir "duraklama" ekle. Bir sonraki adım:
\begin{choices}
  \correctchoice {Eğer dâhilî numara sadece tüm menü oynatıldıktan sonra çevrilebiliyorsa bir "bekleme" ekle.}
  \choice daha önce yapmadıysan, gizli Geliştirici Seçenekleri butonunu görüntülemek için seri numarana 7 kez dokun.
  \choice Ekran görüntüsünü Command ve V tuşlarını basılı tutarak veya Düzenle menüsünden Yapıştır’ı seçerek bir kelime işleme belgesine, bir e-postaya veya bir görüntü düzenleyiciye yapıştır.
\end{choices}
\DashedLineWithText{English Translation}\\
Goal: How to Dial an Extension Number?  
Step: If you’ll enter the extension as soon as the call is answered, insert a “pause.”  
Next step:
\begin{choices}
  \correctchoice {If the extension can only be dialed after the full menu has played, insert a “wait.”}
  \choice If you haven’t done so already, tap your serial number 7 times to reveal the hidden Developer Options button.
  \choice Paste the screenshot into a word processing document, an email, or an image editor by holding down the Command and V keys or selecting Paste from the Edit menu.
\end{choices}
\end{question}

\begin{question}[TurkishPLU / Step Inference]
Hedef: Obsesif Kompulsif Kişilik Bozukluğu Nasıl Tanınır? Örnek Adım:
\begin{choices}
  \choice VPN'in sınırlamalarını bil.
  \choice Hedef SGPT seviyenin ne olduğunu bil.
  \choice SNM’nin prensiplerini benimse.
  \correctchoice {OKKB’nin tanı kriterini bil.}
\end{choices}
\DashedLineWithText{English Translation}\\
Goal: How to Recognize Obsessive-Compulsive Personality Disorder (OCPD)?  
Example Step:
\begin{choices}
  \choice Know the limitations of your VPN.
  \choice Know what your target SGPT level is.
  \choice Adopt the principles of SNM.
  \correctchoice {Know the diagnostic criteria for OCPD.}
\end{choices}
\end{question}

\begin{question}[TurkishPLU / Step Ordering]
Hedef: Tarayıcınızı Güncellemek
\begin{choices}
  \choice Önce: Tarayıcıya uygulanmasını istediğiniz tüm Internet Explorer güncellemelerinin yanına bir onay işareti koyun. Sonra: Internet Explorer için herhangi bir güncelleme olup olmadığını görmek için güncelleme listesini gözden geçirin.
  \correctchoice {Önce: Internet Explorer için herhangi bir güncelleme olup olmadığını görmek için güncelleme listesini gözden geçirin. Sonra: Tarayıcıya uygulanmasını istediğiniz tüm Internet Explorer güncellemelerinin yanına bir onay işareti koyun.}
\end{choices}
\DashedLineWithText{English Translation}\\
Goal: Updating Your Browser  
\begin{choices}
  \choice First: Place a check mark next to all Internet Explorer updates you want to apply to the browser. Then: Review the update list to see if there are any updates for Internet Explorer.
  \correctchoice {First: Review the update list to see if there are any updates for Internet Explorer. Then: Place a check mark next to all Internet Explorer updates you want to apply to the browser.}
\end{choices}
\end{question}

\begin{question}[TurkishMMLU]
220 V gerilimle çalışan ve direnci 484 $\Omega$ olan bir klima günde 5 saat süreyle çalıştırılıyor. Elektrik enerjisinin kWh’i 40 kuruş olduğuna göre klimanın harcadığı 30 günlük enerji bedeli kaç \Lira'dir?
\begin{choices}
  \choice 4
  \choice 5
  \correctchoice 6
  \choice 10
  \choice 15
\end{choices}
\DashedLineWithText{English Translation}\\
A 220 V air conditioner with a resistance of 484 $\Omega$ is operated for 5 hours per day.  
Given that the cost of 1 kWh of electricity is 40 kuruş, what is the energy cost in \Lira{} for 30 days of use?

\begin{choices}
  \choice 4
  \choice 5
  \correctchoice 6
  \choice 10
  \choice 15
\end{choices}
\end{question}

\begin{question}[XCOPA]
Ürün balonlu naylonla paketlenmişti bu yüzden
\begin{choices}
  \correctchoice{kırılgandı.}
  \choice küçüktü.
\end{choices}
\DashedLineWithText{English Translation}\\
Sentence: The product was packaged with bubble wrap, so
\begin{choices}
  \correctchoice{it was fragile.}
  \choice it was small.
\end{choices}
\end{question}

\begin{question}[XQUAD]
Kaynak: Akademi Ödülü kazananı Marlee Matlin Amerikan İşaret Dili(ASL) çevirisini yaparken altı kez Grammy kazanan ve Akademi Ödülü adayı Lady Gaga ulusal marşı söylemiştir.\\\\
Soru: Lady Gaga kaç Grammy kazanmıştır?\\\\
Cevap: \textcolor{ForestGreen}{altı}\\\\
\DashedLineWithText{English Translation}\\
Source: Academy Award winner Marlee Matlin performed the American Sign Language (ASL) interpretation while six-time Grammy winner and Academy Award nominee Lady Gaga sang the national anthem.
\\\\
Question: How many Grammys has Lady Gaga won?
\\\\
Answer:
\textcolor{ForestGreen}{six}
\end{question}

\begin{question}[GECTurk]
Verilen cumlenin yazım hatalarını duzeltin.\\Hatalı Cümle: Büyük yıldızlar transfer ederek, ya'da büyük hocalar getirerek hokus pokus başarıların gelmediği gerçeğinin farkına vardılar.\\\\
Düzeltilmiş hali:
\textcolor{ForestGreen}{Büyük yıldızlar transfer ederek, ya da büyük hocalar getirerek hokus pokus başarıların gelmediği gerçeğinin farkına vardılar.}\\
\\\DashedLineWithText{English Explanation}
In Turkish, the coordinating conjunction "ya da" ("or") is always written as two separate words without an apostrophe.
\end{question}

\begin{question}[MLSum]
Başlık: İzmir'deki orman yangınının tehdit ettiği oteller tahliye edildi\\\\
Metin: İZMİR'in Menderes ilçesindeki tatil beldesi Özdere'de otluk alanda yangın çıktı. Dumanların etkilediği iki otel ise boşaltıldı. İzmir'in tatil beldelerinden Özdere Cumhuriyet Mahallesindeki otluk alanda yangın çıktı. Çıkan yangını söndürmek için 38 arazöz ve 4 dozer aralıksız olarak çalışmalarını sürdürüyor. Öte yandan dumanların etkilediği iki otel boşaltıldı. Helikopter kullanılamıyor Menderes Belediyesi tarafından yapılan açıklamada, "Menderes Belediyesi ekipleri olarak ilk müdahaleyi gerçekleştirdik. İzmir Büyükşehir Belediyemizle de irtibata geçerek İZSU ve İzmir itfaiyemiz hemen yangın alanında müdahaleye başladılar. İzmir Orman Bölge Müdürlüğümüze bağlı ekipler de yangın söndürme çalışmalarını gerçekleştiriyorlar. Havanın karanlık olması nedeniyle uçak ve helikopter ile yangına müdahale gerçekleştirilemiyor. Şu an yol trafiğe kapatılmış durumda. Gümüldür yolu üzerinden ve Ahmetbeyli yolu üzerinden araç trafiği verilmekte. Rüzgarın etkisi çalışmaları zorlaştırsa da ekipler canla, başla yangını kontrol altına almaya çalışıyorlar. Umarız en kısa sürede yangını kontrol altına alabiliriz. Ne yazık ki milli servetimiz, ciğerlerimiz yanıyor. Bir an önce bu felaketin son bulması için canla, başla çalışıyoruz" denildi.\\\\
Özet:\\
\textcolor{ForestGreen}{İZMİR'in Menderes ilçesi Özdere bölgesindeki ormanlık alanda elektrik tellerinin sürtünmesi sonucu orman yangını çıktı. Alevlerin tehdit ettiği bölgede bulunan otellerde konaklayanlar, ekipler tarafından tahliye edildi.}\\
\\\DashedLineWithText{English Translation}\\
Title: Hotels threatened by the forest fire in İzmir evacuated\\\\
Text: A fire broke out in a grassy area in Özdere, a holiday resort in the Menderes district of İzmir. Smoke affected two hotels, which were evacuated. To extinguish the fire, 38 fire trucks and 4 bulldozers have been working nonstop. Menderes Municipality teams performed the first intervention, and İzmir Metropolitan Municipality’s water and fire departments joined the efforts. Teams from the İzmir Forestry Regional Directorate are also fighting the blaze. Due to darkness, aircraft and helicopters cannot be used. Roads are closed to traffic, with vehicles rerouted via the Gümüldür and Ahmetbeyli roads. Despite challenging winds, teams are striving to bring the fire under control. “Our nation’s natural treasure is ablaze.” officials said, urging that the disaster end as soon as possible.\\\\
Summary: \\ 
\textcolor{ForestGreen}{A fire broke out in a grassy area in Özdere, Menderes district of İzmir. Guests at hotels threatened by the flames were evacuated by response teams.}  
\end{question}

\begin{question}[XLSum]
Başlık: 'İklim değişikliği penguenleri tehdit ediyor'\\\\
Metin: ABD'li, İngiliz ve Hollandalı araştırmacılar tarafından yürütülen ve iklim değişikliğinin penguenler üzerindeki etkisini konu alan çalışma, "Nature Climate Change" adlı bilimsel dergide yayımlandı. Makalede, "Büyük penguen" olarak da anılan ve Antartika'da yaşayan bu kuş türüne yönelik asıl tehdidin deniz-buz oranındaki değişimden iddia edildi. Buna göre Antartika'daki buz ve su oranı değişirse, penguenlerin çoğalmaları ve beslenmeleri olumsuz etkilenecek. Çalışma, penguen grupları arasında farklı dinamiklerin etkili olacağını ancak yine de tüm gruplarda sayının azalacağını savunuyor. Araştırmacılar, devletlerin penguenleri "nesli tükenmekte olan kuşlar" olarak korumaya alması önerisinde bulundu. Ancak korumaya yönelik tedbirler, turizm ve balıkçılık alanında kısıtlamalara neden olabiliyor. 'Neslinin tükenmesi tehdidi var' Çalışmayı yürüten ekibin başında "Woods Hole Oceanographic" Enstitüsü'nden Stephanie Jenouvrier yer alıyor. Doktor Jenouvrier, tüm penguen nüfusunun yüzde 19 ile 33 arasında bir oranda azalacağını belirtiyor. Jenouvrier penguenlerin "Yakın bir gelecekte önemli oranda nüfusunu kaybedeceğini ve muhtemelen neslinin tükenmesi tehlikesiyle karşı karşıya kalacağını" söyledi. Antarktika'daki Ross denizi çevresinde yaşayan penguen gruplarının iklim değişikliğinden en son etkilenenler olacağını kaydeden Jenouvrier'a göre bunun sebebi, bölgedeki deniz ve buz dağılımının penguenler için hala elverişli olması. Jenouvrier sözlerine şöyle devam etti: "Ross denizinde penguenlerin yaşadığı bölgenin korunması ve geliştirilmesi, nesil tükenmesi tehdidine karşı zaman kazandıracaktır. Böylece sera gazının azaltılması konusunda gerekli müzakereler yapılabilecek, stratejiler belirlenebilecektir." Aylarca yol kat ediyor, yemek arıyorlar Penguenler, yavrularını beslemek için aylarca yuvadan uzak, yemek arıyorlar. Antarktika buzulları boyunca uzun müsefaler kat eden penguenler, denize eriştikleri yerlerden karides gibi yiyecekler topluyorlar. Penguenler yemek ararken yırtıcı hayvanlardan korunmak gibi çeşitli nedenlerle ideal miktarda buzul tabakaya ihtiyaç duyuyor. Buzul ve deniz miktarındaki değişimin penguenlerin beslendiği karides gibi canlıların verimliliğini de etkileyeceği belirtiliyor. Penguenlerin ana besin kaynağı olan karides ve benzeri deniz kabuklularının üremesinin, buzul deniz dağılımından etkilendiği ifade ediliyor. Buzulların artması karides ve diğer kabuklular için olumlu olarak değerlendiriliyor. Ancak bu durum, penguenlerin denize ulaşmak için daha uzun mesafe kat etmesi anlamına geliyor. Uydudan yapılan ölçümlerde, Antartika'da buzul-su seviyesinin daha önce görülmeyen bir seviyeye yükseldiği görülüyor. Ancak iklim modelleme yazılımları, bu durumun ileride tersine döneceğini belirtiyor.\\\\
Özet:\\
\textcolor{ForestGreen}{İklim değişikliğinin Antarktika'daki penguen nüfusunu olumsuz etkileyebileceği belirtiliyor. Yapılan bir çalışmaya göre, sayıları 600 bini bulan penguenlerin 2100 yılı itibariyle beşte biri oranında azalabileceği ifade ediliyor.}
\DashedLineWithText{English Translation}\\
Title: 'Climate change threatens penguins'\\\\
Text: A study conducted by US, British, and Dutch researchers on the impact of climate change on penguins was published in the scientific journal "Nature Climate Change." The article states that the main threat to this bird species, also known as the “Emperor penguin” living in Antarctica, is claimed to be changes in the sea–ice ratio. According to this, if the ratio of ice to water in Antarctica changes, penguins’ breeding and feeding will be negatively affected. The study argues that different dynamics will be at play among penguin colonies, but that numbers will decline in all groups. The researchers recommended that governments list penguins as “endangered birds.” However, conservation measures can lead to restrictions in tourism and fishing.  
Under the leadership of Stephanie Jenouvrier from the Woods Hole Oceanographic Institution, the team determined that the total penguin population will decrease by between 19\% and 33\%. Jenouvrier said that penguins will lose a significant portion of their population in the near future and may face the risk of extinction. According to Jenouvrier, the colonies living around the Ross Sea in Antarctica will be the last to be affected by climate change, because the distribution of sea and ice in that region remains favorable for penguins. Jenouvrier continued: “Protecting and enhancing the region where penguins live in the Ross Sea will buy time against the threat of extinction. This will allow for necessary negotiations and strategies to reduce greenhouse gases to be developed.”  
Penguins travel for months and search for food to feed their chicks. Penguins traverse long distances along Antarctic ice, collecting krill and similar prey from sea access points. Penguins need an ideal amount of ice shelf for various reasons, such as protection from predators while foraging. Changes in ice and sea levels will also affect the productivity of krill and other crustaceans that penguins eat. An increase in ice benefits krill and other crustaceans, but forces penguins to travel longer distances to reach the sea. Satellite measurements show that the ice–water ratio in Antarctica has reached unprecedented levels. However, climate models predict this trend will reverse in the future.\\\\
Summary:\\
\textcolor{ForestGreen}{It is stated that climate change could negatively affect the penguin population in Antarctica. According to a study, the population of around 600,000 penguins could decrease by one fifth by 2100.}  

\end{question}

\begin{question}[WikiHowSum]
Metin: Yatabileceğin en doğal uyku pozisyonunda uzan. Birşey tutma, bacaklarını yatakta tut, başını kaldırma. Eğer normalda sırt üstü uyuyorsan numaradan uyurken de öyle yap. Böylece seni tanıyan insanlar şüphelenmez. Doğal uykunda çok az hareket edersin. Gerçekten uyuyormuş izlenimi yaratmak için en iyisi hiç hareket etmemek. Biri seni uzun bir süre boyunca izlemediği sürece hareket etmen beklenmez. Göz kapaklarını fazlaca sıkarak kapatmaktan kaçın. En iyi uyuyor izlenimi için, göz kapakların dâhil tüm kaslarının rahat olmalı.  Gözlerini kapattıktan sonra göz kapaklarının kırpışmasını engellemek için aşağı doğru bak. Uyurken gözlerin her zaman tam kapalı olmaz. Göz kapaklarının düşerek nazikçe kapanmasına izin ver; hâlâ göz kapaklarının arasından etrafı biraz görebilirsin. Yavaş, hatta derin nefesler al. Nefes almanı rahatlatmalı ve mümkün olduğunca eşit aralıklarla nefes alıp vermelisin. Nefes alırken kafandan sayıp, aynı sürede vermeye çalış. Bunu her nefesinde yap. Eğer yüksek bir ses duyarsan ya da biri sana dokunursa kısa ve ani bir nefes al ve vücudunu hafifçe titret. Uyurken bile, vücutlarımız etrafımızda olan şeylerin farkındadır. Sahte uykunu, odadaki seslere ve hareketlere bilinçsiz görünen tepkiler ekleyerek sat.  Rahatsızlığa tepki verdikten sonra, vücudunun gevşemesine ve nefesinin yavaş ve dengeli bir duruma dönmesine izin ver. Sakın gülümseme ve gözlerini açma, yoksa aslında uyanık olduğun hemen anlaşılır.\\\\
Özet:
\textcolor{ForestGreen}{Doğal bir uyku pozisyonu seç. Yatakta hareketsiz bir şekilde yat. Gözlerini nazikçe kapat. Ritmik bir şekilde nefes alıp ver. Seslere ve dokunmaya tepki ver.}
\\\DashedLineWithText{English Translation}
\\
Text: Lie in the most natural sleep position possible. Don’t hold anything, keep your legs on the bed, don’t lift your head. If you normally sleep on your back, do so here as well. That way, people who know you won’t suspect. You move very little in natural sleep. To create the impression of truly sleeping, it’s best not to move at all. As long as no one watches you for a long time, movement isn’t expected. Avoid squeezing your eyelids tightly shut. For the best sleeping impression, all your muscles—including your eyelids—should be relaxed. After closing your eyes, look downward to prevent eyelid twitching. Eyes are never fully closed when sleeping. Allow your eyelids to fall gently and close; you can still see a little through them. Breathe slowly, even deeply. Your breathing should be relaxed and at as equal intervals as possible. Count in your head as you inhale, and try to exhale in the same time. Do this with each breath. If you hear a loud noise or someone touches you, take a short, sudden breath and slightly shake your body. Even during sleep, our bodies are aware of things around us. Sell your fake sleep by adding unconscious-seeming reactions to sounds and movements in the room. After reacting, allow your body to relax and your breathing to return to a slow, balanced state. Never smile or open your eyes, or it will immediately reveal you are actually awake.

Summary:
\textcolor{ForestGreen}{Choose a natural sleep position. Lie motionless on the bed. Gently close your eyes. Breathe rhythmically. Respond to sounds and touch.}
\end{question}

\begin{question}[WMT16\textsubscript{EN-TR}]
Translate English to Turkish.\\\\
English: Norway's rakfisk: Is this the world's smelliest fish?\\
Turkish:
\textcolor{ForestGreen}{Norveç'in rakfisk'i: Dünyanın en kokulu balığı bu mu?}
\end{question}

%% file: sections/lira.tex
\newcommand{\Lira}{%
\begin{tikzpicture}[x=0.08em, y=0.08em, xscale=0.03, yscale=-0.03, inner sep=0pt, outer sep=0pt]
\fill (54.3355,9.3092) .. controls (70.3869,9.3092) and (79.7110,9.3092) ..
  (82.3075,9.3092) .. controls (82.3075,12.1418) and (82.3075,24.2985) ..
  (82.3075,45.7791) .. controls (82.3075,67.2598) and (82.3075,79.2984) ..
  (82.3075,81.8950) -- (167.2860,51.0903) .. controls (167.2860,59.3521) and
  (167.2860,66.7877) .. (167.2860,73.3971) -- (82.3075,104.2018) .. controls
  (82.3075,105.1460) and (82.3075,107.1525) .. (82.3075,110.2211) .. controls
  (82.3075,113.2898) and (82.3075,115.2962) .. (82.3075,116.2404) --
  (128.3375,99.9529) -- (167.2860,85.4358) .. controls (167.2860,86.8521) and
  (167.2860,90.4518) .. (167.2860,96.2351) .. controls (167.2860,102.0184) and
  (167.2860,105.5001) .. (167.2860,106.6804) .. controls (167.2860,107.6246) and
  (167.0499,108.0967) .. (166.5778,108.0967) -- (84.7861,137.4851) .. controls
  (83.8419,137.9572) and (83.0157,138.4293) .. (82.3075,138.9014) .. controls
  (82.3075,153.3005) and (82.3075,173.1878) .. (82.3075,198.5633) .. controls
  (82.3075,223.9388) and (82.3075,243.8262) .. (82.3075,258.2253) .. controls
  (82.3075,258.2253) and (82.3075,258.3433) .. (82.3075,258.5794) .. controls
  (82.3075,258.8154) and (82.3075,258.9334) .. (82.3075,258.9334) .. controls
  (103.7882,256.3369) and (122.7903,248.1931) .. (139.3139,234.5021) .. controls
  (157.4899,219.6309) and (169.6465,201.1009) .. (175.7838,178.9121) .. controls
  (178.3804,168.7619) and (179.6787,158.6117) .. (179.6787,148.4614) .. controls
  (179.6787,148.4614) and (189.1207,148.4614) .. (208.0048,148.4614) .. controls
  (208.0048,171.3584) and (202.4576,192.9571) .. (191.3632,213.2575) .. controls
  (183.5735,228.8369) and (172.9512,242.2918) .. (159.4963,253.6223) .. controls
  (146.9856,264.7167) and (132.9405,273.2145) .. (117.3611,279.1158) .. controls
  (97.0607,286.9055) and (76.0522,289.6201) .. (54.3355,287.2596) .. controls
  (54.3355,271.9163) and (54.3355,248.9013) .. (54.3355,218.2146) .. controls
  (54.3355,187.5279) and (54.3355,164.3949) .. (54.3355,148.8155) .. controls
  (54.3355,148.8155) and (52.8011,149.4057) .. (49.7325,150.5859) --
  (1.2239,167.9357) .. controls (1.2239,160.8541) and (1.2239,153.4185) ..
  (1.2239,145.6288) -- (54.3355,126.5087) .. controls (54.3355,125.0924) and
  (54.3355,120.9615) .. (54.3355,114.1160) .. controls (51.7389,115.2962) and
  (48.2571,116.7126) .. (43.8902,118.3649) .. controls (39.5232,120.0173) and
  (36.7496,120.9615) .. (35.5694,121.1975) -- (7.5973,131.4658) .. controls
  (7.5973,131.4658) and (6.8301,131.7018) .. (5.2958,132.1739) .. controls
  (3.7615,132.6460) and (2.4042,133.0001) .. (1.2239,133.2361) .. controls
  (1.2239,121.9057) and (1.2239,114.4701) .. (1.2239,110.9293) --
  (54.3355,92.1632) .. controls (54.3355,81.7770) and (54.3355,67.9680) ..
  (54.3355,50.7362) .. controls (54.3355,33.5045) and (54.3355,19.6955) ..
  (54.3355,9.3092) -- cycle;

\end{tikzpicture}}

%% file: tables/datasets.tex
\renewcommand{\arraystretch}{1.3}{
\begin{table*}[ht!]
\centering
\resizebox{\textwidth}{!}{%
\begin{tabular}{ c|l|l|l|r|c}
\multicolumn{6}{c}{\textit{Language Understanding Tasks}} \\ 
\hline
Task & Dataset & Source / Domain & License & \#examp. & Metric \\
\hline
\multirow{3}{11em}{\centering Extractive \\ Question Answering} & XQUAD & Wikipedia & CC BY‑SA 4.0 & 1190 &\multirow{3}{5em}{Exact Match} \\ \cline{2-5}
& TQUAD & Wikipedia & MIT & 892 &\\ \cline{2-5}
& MKQA & Wikipedia & CC BY‑SA 3.0 & 10000\\ \hline
\multirow{8}{11em}{\centering Multiple Choice \\ Question Answering} & Exams & High School Exams & CC-BY-SA 4.0 & 393 & \multirow{8}{5em}{Accuracy} \\ \cline{2-5}
& Belebele & Online Web Pages & CC-BY-SA 4.0 & 900 &\\  \cline{2-5} 
& TurkishPLU & WikiHow & CC BY‑NC‑SA 3.0 & 3124 &\\  \cline{2-5} 
& XCOPA & Crowd Sourcing & CC-BY-SA 4.0 & 600 &\\  \cline{2-5}
& TurkishMMLU & High School Exams & - & 900 &\\  \cline{2-5} 
& Proverbs & Turkish Dictionaries & GPL-3.0 & 1730 &\\  \cline{2-5} 
& BilmeceBench & Turkish riddles & MIT & 442 &\\ \cline{2-5} 
& CircumflexTR & Turkish Dictionaries & MIT & 72 &\\ \hline
\multirow{4}{11em}{\centering Text Classification} & IronyTR & Twitter & - & 600 & \multirow{4}{5em}{Accuracy} \\ \cline{2-5}
& NewsCat & Online Newspapers & - & 250 &\\ \cline{2-5}
& OffensEval & Twitter & CC BY-SA 2.0  & 3528 &\\ \cline{2-5}
& STSb & Multi-Domain & CC BY-SA 4.0  & 1389 &\\ \hline
\multirow{3}{11em}{\centering Natural Language Inference} & XNLI & SNLI \& MNLI & CC BY 4.0 & 5010 & \multirow{3}{5em}{Accuracy} \\ \cline{2-5}
& MNLI & Crowd Sourcing & CC BY-SA 3.0  & 10000 &\\ \cline{2-5}
& SNLI & Image Captions & CC BY‑SA 4.0 & 10000 &\\ \hline
\multicolumn{6}{c}{\textit{Language Generation Tasks}} \\ 
\hline
Task & Dataset & Source / Domain & License & \#examp. & Metric \\
\hline
\multirow{3}{11em}{\centering Summarization} & MLSum & Online Newspapers & Other &  12775 & \multirow{3}{5em}{ROUGE-2} \\ \cline{2-5}
& XLSum & BBC News & CC BY‑NC‑SA 4.0 & 3400 &\\ \cline{2-5}
& WikiLingua & WikiHow & CC BY‑NC‑SA 3.0 & 3000 &\\ \hline
Machine Translation & WMT16\textsubscript{EN-TR} & Online Newspapers & - & 3000 & BLEU-4 \\ \hline
Grammatical Correction & GECTurk & Newspapers & Apache-2.0 & 20800 & Exact Match \\ \hline
\end{tabular}
}%
\caption{\label{table:datasets}
    The overview of the tasks \& datasets included in the \pergel~benchmark. Most of the listed datasets are the Turkish splits of a larger multilingual dataset with few exceptions which are TurkishPLU, TQUAD, IronyTR, NewsCat and GECTurk.
}
\end{table*}
}